\documentclass[preprint,review,10pt]{elsarticle}

\usepackage{amssymb}
\usepackage{amsmath}
\usepackage{subcaption}
\usepackage{todonotes}
\usepackage{algorithm}
\usepackage{algorithmic}
\usepackage{adjustbox}
\usepackage{hyperref}
\usepackage[capitalize]{cleveref}
\newcommand{\red}[1]{\textbf{\textcolor{red}{#1}}}

\journal{Computers \& Operations Research}

\begin{document}
\begin{frontmatter}


\cortext[corr1]{Corresponding author}

\title{Automated Placement of Analog Integrated Circuits using Priority-based Constructive Heuristic}

\affiliation[dcefel]{organization={DCE, FEE, Czech Technical University in Prague}, addressline={Technická 2}, city={Praha}, postcode={160 00}, country={Czech Republic}}
\affiliation[ciirc]{organization={IID, CIIRC, Czech Technical University in Prague}, addressline={Jugoslávských partyzánů 1580/3}, city={Praha}, postcode={160 00}, country={Czech Republic},}

\author[dcefel,ciirc]{Josef Grus\corref{corr1}}
\ead{josef.grus@cvut.cz}
\author[ciirc]{Zdeněk Hanzálek}
\ead{zdenek.hanzalek@cvut.cz}

\begin{abstract}
This paper presents a heuristic approach for solving the placement of Analog and Mixed-Signal Integrated Circuits. Placement is a crucial step in the physical design of integrated circuits. During this step, designers choose the position and variant of each circuit device. We focus on the specific class of analog placement, which requires so-called pockets, their possible merging, and parametrizable minimum distances between devices, which are features mostly omitted in recent research and literature. We formulate the problem using Integer Linear Programming and propose a priority-based constructive heuristic inspired by algorithms for the Facility Layout Problem. Our solution minimizes the perimeter of the circuit's bounding box and the approximated wire length. Multiple variants of the devices with different dimensions are considered. Furthermore, we model constraints crucial for the placement problem, such as symmetry groups and blockage areas. Our outlined improvements make the heuristic suitable to handle complex rules of placement. With a search guided either by a Genetic Algorithm or a Covariance Matrix Adaptation Evolution Strategy, we show the quality of the proposed method on both synthetically generated and real-life industrial instances accompanied by manually created designs. Furthermore, we apply reinforcement learning to control the hyper-parameters of the genetic algorithm. Synthetic instances with more than 200 devices demonstrate that our method can tackle problems more complex than typical industry examples. We also compare our method with results achieved by contemporary state-of-the-art methods on the MCNC and GSRC datasets.
\end{abstract}


\end{frontmatter}





\section{Introduction} 
\label{sec:introduction}
\subsection{Motivation}
Most Integrated Circuits (ICs) today consist of analog and digital components. Their physical design is highly complex, but the sources of complexity differ between analog and digital ICs \citep{survey}. Digital designers have to deal with a large number of rectangular devices, but all the devices share the same height and are placed in rows rather than freely, which resembles 1D Bin Packing Problem \citep{survey_1dbpp}. On the other hand, Analog and Mixed-Signal (AMS) ICs usually contain fewer devices, which may have different sizes and voltage levels and can be freely placed on the canvas. As a result, designers must consider a complex set of rules and constraints to mitigate the negative effects of noise and process variations. The lack of automation tools means that most of the work is still done manually, leading to a time-consuming and error-prone workflow.

\subsection{Outline}
The physical design, or layout, is usually divided into placement and routing. Each rectangular device is assigned its position and orientation during the placement, and interconnections between connected devices are determined during the routing phase. In this paper, we specifically deal with the placement phase. Thanks to the rectangular shapes of the devices, we tackle the problem as an extension of the rectangle packing. Usually, designers want to minimize the total area and approximated wire length of the IC, and they need to satisfy common requirements such as proximity constraints (e.g., various minimum distances between devices) and symmetry constraints. Additional constraints depend on the chosen IC technology. In order to support influential BCD technology (combines analog, digital, and high-voltage components), we also need to model so-called pockets - an additional empty space among the devices that can be merged when the devices have the same voltage levels; this leads to more compact designs, but it adds another level of complexity to the already complicated combinatorial problem.

In this paper, we formulate and solve the problem of the placement of AMS ICs using combinatorial optimization methods. The rest of the paper is organized as follows. The previous works related to the placement of ICs are presented in \cref{sec:rel_work}. In \cref{sec:ilp_approach}, the placement problem is formalized, and the Integer Linear Programming (ILP) model is formulated. In \cref{sec:heur_const}, a constructive heuristic is proposed. We propose to use metaheuristics Genetic Algorithm (GA) and Covariance Matrix Adaptation Evolution Strategy (CMA-ES) to perform the search. We also describe local search approaches to further improve the found placement. In \cref{sec:rl_control_text}, we describe our approach of applying reinforcement learning for parameter control of GA. In \cref{sec:experiments}, results are presented, including a comparison of automated and manual placements provided by industry partner STMicroelectronics, and finally, in \cref{sec:conclusion}, conclusions are drawn, discussing the applicability of this approach.

\section{Related Work}
\label{sec:rel_work}
\subsection{Literature overview}
The placement phase of the physical design is a problem of placing a given set of devices, described by the netlist input file and represented by rectangles, in such a way that all the design rules are satisfied. Designers usually try to minimize the chip's area and the estimated wire length.

There are two main categories of problem representation prevalent in literature. The first category encodes the solution using topological representation, which determines the relative positions between devices. The representation commonly used in the literature is sequence-pairs, originally proposed in \citep{murata_sequence_pairs}. It utilizes two permutations of devices, which encode the placement. Authors of \citep{sequence_pairs} extend this representation to handle crucial features such as symmetry groups. The sequence pair representation was further extended to handle general constraints such as abutment (two devices are placed exactly next to each other)  in \citep{handling}, and this work finally became the core of the placer of the open-source tool ALIGN \citep{align}. In \citep{pisinger_seqforstrip}, an efficient algorithm for transforming the representation to the actual packing was found. 

Another topological representation is B*-trees. Used in \citep{symmetry_island_mcnc,gsrc_dp,gsrc_memetic}, the solution is represented by a binary tree, with each device stored in a node. Simulated Annealing (SA) is often used to guide the search with topological representations. On the other hand, the B*-trees-based approach in \citep{representation_as_trees} is deterministic. The method exploits the hierarchical structure of the ICs, which guides the bottom-up enumeration of the parts of the circuit.

The second category of problem representation is absolute representation, which describes each device by its coordinates, i.e., absolute position. Thus, symmetry and other constraints can be easily formulated. Nevertheless, placements with illegal overlaps are part of the search space and must be dealt with. This approach was used early in \citep{koan_anagram}. Experimentally chosen costs of the criterion function lead to feasible placements comparable to high-quality manual designs. More recent work of \citep{camose_mcnc} utilizes the constrained multi-objective metaheuristic. Other absolute representation approaches \citep{magical,date_placer_fdgd} often build on works similar to \citep{ntuplace}. Solution of \citep{ntuplace} firstly determines the global placement, where illegal overlaps are allowed, and then the feasible solution is obtained by legalization. In \citep{gaafteropt}, different manufacturing layers were considered, so the devices that do not share the same layer can arbitrarily overlap. 

While the achievements of the mentioned approaches cannot be underestimated, we cannot directly apply their results to our problem. The main reasons are various minimum allowed distances and pockets. Such features are critical for BCD technology and are only partly covered, e.g., in \citep{gaafteropt}. However, we can compare the performance of our approach with both topological and absolute representation methods on instances that do not rely on such constraints.

ILP was previously used to solve the placement problem. The formulation essentially combines the advantages of both previously mentioned approaches. The authors of \citep{hiearchichal_mcnc} utilized hierarchical decomposition of the problem to achieve an acceptable computation time. However, the ILP solvers usually struggle with larger instances, even when using decomposition and other techniques. A novel self-organizing approach was outlined in \citep{swarm}. An end-to-end pipeline \citep{graph_placement} used reinforcement learning to place macros of IC one by one.

Due to their similarity, we also mention rectangle packing results - we can interpret the placement problem as an extension of rectangle packing with connectivity and other features. Exact approaches for 2D rectangle packing are known from \citep{CSP_ortho,CSP_abs_rela}, where ILP and Constraint Programming (CP) models were shown. The CP search was strengthened with domain-specific branching. There is a plethora of efficient constructive and improvement heuristics thoroughly researched for rectangle and strip packing \citep{packing_heur_survey}. Bottom-left heuristics were studied in \citep{Hopper1999AGA,Crainic2008ExtremePH}. The fitness-based packing heuristic was paired with SA for a 2-D knapsack packing in \citep{SA_hybrid_packing}. Finally, authors of \citep{grasp_packing} relied on the GRASP metaheuristic to solve strip packing. However, due to minimum allowed distances, symmetry groups, and connectivity, we cannot simply apply the previous results in the domain of AMS IC placement.

Facility Layout Problem (FLP) is especially noteworthy since its goal is to assign positions of the machines within the factory and to minimize the travel distance between related machines. Exact solutions using custom branch and bound \citep{fac2} or general ILP solver \citep{opt_fac_mip} were successful. The latter work even incorporated the aisle design directly into the layout phase. In \citep{kubalik,kubalik2}, FLP was solved using GA with a priority-based constructive heuristic, considering different variants for each facility. An evolutionary approach for solving FLP was presented in \citep{fac_mip_but_nonexact}. The ILP solver solved a sequence of increasingly more complex models derived from the actual problem instance until the final solution was found.

\subsection{Contributions}
While the research on automation of the placement of the AMS ICs has greatly improved in recent years, there are still areas that need to be thoroughly investigated. BCD technology still needs to be addressed since it requires various minimum allowed distances between devices, isolated pockets, and their merging. These features are critical, yet mostly omitted in literature (merging was mentioned in \citep{koan_anagram}, and a similar concept was shown in \citep{gaafteropt}).

We build on our previous conference papers \citep{icores23,icores2023-rozsireni}, and FLP \citep{kubalik}. We re-use our conference paper problem statement and ILP formulation with Force-Directed Graph Drawing (FDGD) warm start \citep{icores23} as a baseline solution for comparison, and we tackle the placement problem using the priority-based constructive heuristic and metaheuristics.

We summarize the main contributions of this paper as follows:
\begin{itemize}
    \item Solution to an industry-relevant placement problem formulated in \citep{icores23}. To the best of our knowledge, this problem with associated constraints has not been addressed before. Pockets, variants, and minimum allowed distances are all considered.
    \item Priority-based constructive heuristic inspired by \citep{kubalik} maps each indirect representation to a feasible placement. Search for a high-quality solution is guided by the GA and CMA-ES metaheuristics. The method can solve instances with around 200 devices, improving the results of \citep{icores23}. 
    \item Proposal and evaluation of reinforcement learning parameter control for GA above, which dynamically modifies the GA's hyper-parameters controlling the selection of parents and crossover.
\end{itemize}

We evaluate our solution on synthetically generated instances, which we use to compare the heuristic approach with ILP results of \citep{icores23}. We also compare our solution with the placer of ALIGN \citep{align}, and with relatively recent papers solving MCNC and GSRC benchmarks, which highlight the competitiveness of our solution. Finally, we compare placements found by our algorithm with manual designs provided by industry partner STMicroelectronics.

\section{Analog Placement Model}
\label{sec:ilp_approach}
\subsection{Problem formulation}
\begin{figure}
        \centering
        \includegraphics[width=0.55\textwidth]{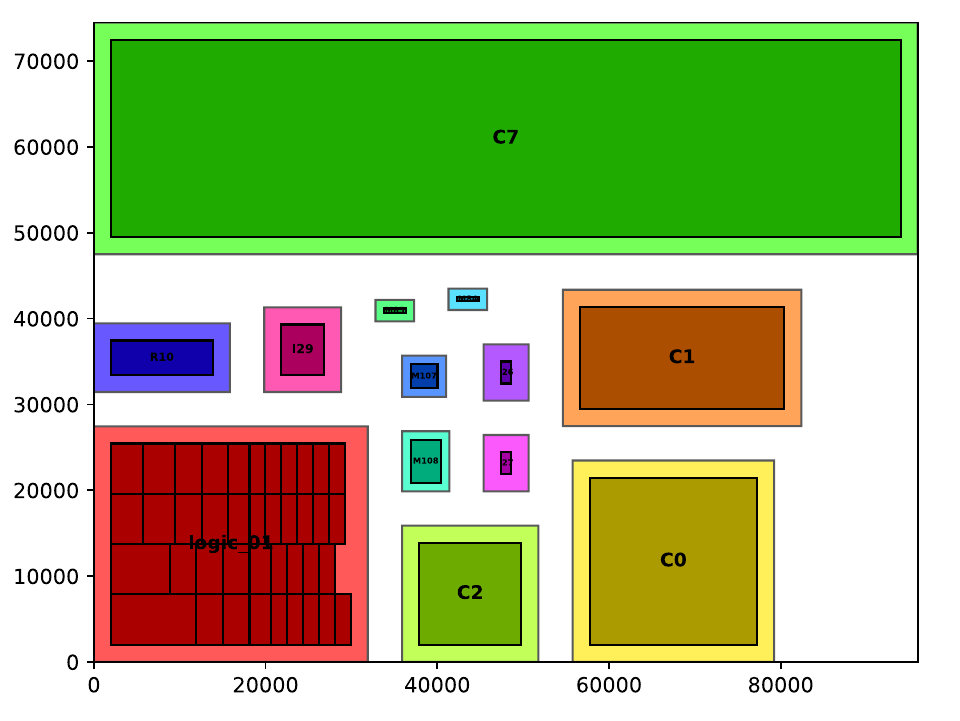}
        \caption{Illustrative example of automatically generated placement of real-life netlist, with the increased minimum allowed distances to simplify the visualization.}
        \label{fig:ill_opa}
\end{figure}

The placement problem assigns the exact positions and orientations to the devices (transistors, resistors, etc.) described by the provided netlist so the devices do not overlap and the placement area and connectivity are minimized. In addition to single devices, there are higher-level design blocks called topological structures, such as current mirrors and differential pairs. They consist of devices that need to be packed in a regular pattern to function properly. Finally, placement constraints and technological requirements are provided as well. Example placement created from real-life instance provided by industry partner STMicroelectronics is shown in \cref{fig:ill_opa}. 

Each device or topological structure is modeled and further denoted as a rectangle and is associated with its set of width-and-height variants $R_i$. These variants refer simply to rotation in the case of a single device or multiple different aspect ratio configurations in the case of topological structures. The sizes of the configurations depend on the arrangement of the structure's internal devices. Arbitrary minimum allowed distance, or spacing, is defined between each pair of rectangles depending on the chosen technology process and the designer's requirements.

The concept of pockets - additional free space around the devices - also needs to be considered to accommodate the BCD process. Pockets and additional spacing ensure that devices with different voltage levels do not negatively affect each other. However, the pockets of devices with the same voltage level can be merged if the designer allows it; this concept is called pocket merging.

Additionally, several devices can be part of the symmetry group - each symmetry pair and self-symmetric device of the group needs to respect their common axis of symmetry. Blockage areas restrict the placement of specific devices within a specific part of the canvas. Furthermore, pairs of devices that should be placed as close to or, conversely, as far from each other as possible can be defined as well. Finally, we also model connectivity with external components so the optimized placement will fit with the other parts of the overall design.

\subsection{ILP model}
\label{sec:ilp}
\subsubsection{Main constraints and criteria}
The placement model was formulated using ILP in our previous work \citep{icores23}, which we extended to include additional features. Therefore, we could use a general ILP solver as a baseline for comparison.

\begin{figure}
	\centering
	\includegraphics[width=.65\linewidth]{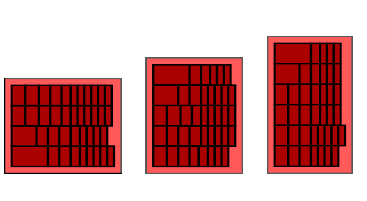}
	\caption{Examples of different variants produced by scheduling-based enumeration \citep{lpt}, with internal devices shown in a darker color and outer pocket encompassing them. The left variant is used in \cref{fig:ill_opa}.}
	\label{fig:lpt_example}
\end{figure}

Let $\mathcal{I} = \lbrace 1, \dots, n \rbrace$ be a set of indices of rectangles. Each rectangle is represented by coordinates variables of its bottom-left corner $(x_i,y_i)$ and chosen width-and-height variant $(w_i, h_i) \in R_i$, $|\,R_i\,| = m_i$. When the pocket of the device is considered, the width and height of the rectangle's variants are appropriately enlarged. Variants are selected using binary variables $s_i^k$; $k$-th variant is selected if $s_i^k=1$. Different variants of each structure are exhaustively enumerated beforehand, either using matrix-array pattern enumeration when the internal devices have the same size or scheduling-based enumeration \citep{lpt} when the devices of the structure only share a single dimension. Such variability (i.e., different variants) of rectangles to be placed was previously explored in \citep{swarm}, where the authors also considered different proportions of the primitive devices. Examples of the variants created by the scheduling-based approach are shown in \cref{fig:lpt_example}. 

Variables $W,\,H$ define the width and height of the placement's bounding box. Minimum allowed distances are enforced by binary variables $r_{i,j}^k$ and inequalities \eqref{pos0} - \eqref{rel2_eq}, forming the OR-constraint. At least one of the inequalities, which corresponds to the relationship (left/right/over/under) between rectangles, must be valid ($r_{i,j}^k = 1$) without the big-$M$ term \citep{bigM}. Parameter $a_{i,j}$ defines the minimum allowed distance between rectangles. By appropriately shifting the value of $a_{i,j}$ to negative, pocket merging becomes available. Note that redundant constraints $r_{i,j}^1 + r^3_{i,j} \le 1$ and $r_{i,j}^2 + r^4_{i,j} \le 1$ were incorporated into the model to improve the optimization performance.

\begin{align}
	\label{main}
	& x_i + w_i \le W,~~ y_i + h_i \le H&\forall i \in \mathcal{I}\\
	& \sum_{k=1}^{m_i} s_i^{k} = 1& \forall i \in \mathcal{I}\\
	& w_i = \sum_{k=1}^{m_i} w^k_i \cdot s_i^k,~~h_i = \sum_{k=1}^{m_i} h^k_i \cdot s_i^k &\forall i \in \mathcal{I}\\
	& \sum_{k=1}^{4} r^k_{i,j} \ge 1 &\forall i\forall j \in  \mathcal{I}:~i < j \label{pos0} \\
	& x_i + w_i + a_{i,j}\le x_j + M  (1 - r^1_{i,j}) &\forall i\forall j \in  \mathcal{I}:~i < j \label{pos1} \\ 
	& y_i + h_i + a_{i,j} \le  y_j + M  (1 - r^2_{i,j}) &\forall i\forall j \in  \mathcal{I}:~i < j \\
	& x_j + w_j + a_{i,j} \le x_i + M  (1 - r^3_{i,j}) &\forall i\forall j \in  \mathcal{I}:~i < j \label{pos2} \\ 
	& y_j + h_j + a_{i,j}\le  y_i + M (1 - r^4_{i,j})  &\forall i\forall j \in  \mathcal{I}:~i < j \label{rel2_eq}\\
	& x_i, y_i, w_i, h_i \ge 0 & \forall i \in  \mathcal{I}\\
	& W,~H \ge 0 \\
	& s^k_i \in \lbrace 0, 1 \rbrace & \forall i \in \mathcal{I}~\forall k \leq m_i \\
	\notag& r^k_{i,j} \in \lbrace 0, 1 \rbrace &\forall i\forall j \in  \mathcal{I}:~i < j \\
	&& \forall k \in \lbrace 1,2,3,4\rbrace\label{last}
\end{align}

Symmetry groups consist of symmetry pairs and self-symmetric devices. It is required that both rectangles of the symmetry pair use the same variant and that there is a common (horizontal or vertical) axis of symmetry for the entire group. Consider the symmetry group $G$, with the vertical axis of symmetry, $p$ symmetry pairs, and $q$ self-symmetric rectangles. Each pair or self-symmetric rectangle is described by their indices:
\begin{equation}
G = \lbrace (i^1,j^1), \dots, (i^p, j^p), (i^1,-), \dots, (i^q,-)\rbrace    
\end{equation}

With the variable $x_G \in \mathbb{R}$ encoding the position of the axis of symmetry, the symmetry constraint is enforced using the following equations. Note that the same-variant per pair constraint is enforced by requiring that the widths and heights of the rectangles of the symmetry pair are the same.
\begin{align}
    2 \cdot x_G&= x_{i} + x_{j} + w_{i}   & \forall (i,j) \in G \label{symmetrystart}\\
    y_{i} &= y_{j}& \forall (i,j) \in G\\
    w_{i} &= w_{j} & \forall (i,j) \in G\\
    h_{i} &= h_{j} & \forall (i,j) \in G\\
    2 \cdot x_G &=  2\cdot x_i + w_i&\forall (i,-) \in G \label{symmetryend}
\end{align}

The equations \eqref{symmetrystart}-\eqref{symmetryend} describe the constraints for the symmetry group with the vertical axis of symmetry. In the case of a horizontal symmetry group, the equations would be analogously defined.

Blockage area requirement is handled simply by introducing the dummy rectangles. These dummy rectangles have fixed coordinates; additional binary variables are needed to model the relative position constraints \eqref{pos0} - \eqref{rel2_eq}.

The aspect ratio of the placement, defined by its width and height $W$, $H$, is constrained using the user-defined bounds $l_R$, $u_R$, such that:
\begin{equation}
    0  \le l_R \le R \le u_R \le 1
\end{equation} where $R=\frac{\min \lbrace W, H\rbrace}{\max\lbrace W,H\rbrace}$ is the aspect ratio of the design. Therefore, the following inequalities and additional binary variable $r_R$ are used to handle the non-convex solution space. When $r_R = 0$, we enforce that the width $W$ is less than height $H$ given parameter $u_R$, and vice versa. Without the binary variable, a contradiction $H \le u_R \cdot W \le u_R^2 H$ would occur.
\begin{align}
	& l_R \cdot W \le H  \le u_R \cdot W + M \cdot (1-r_R)  \\
	& l_R \cdot H \le W \le u_R \cdot H + M \cdot r_R\\
	& r_R \in \lbrace 0; 1 \rbrace
\end{align}

To minimize the area defined by expression $W\cdot H$ within the ILP framework, we approximate it using the half perimeter of the placement $\mathcal{L}_{area}=W+H$. This approximates the area well, when the aspect ratio is close to 1.

Nets describe connectivity between the devices. Each net $e$ of the set of nets $E$ is associated with its set of connected devices $L_e$ and its cost $c_e > 0$ (defined by the user or set to 1). Net can be omitted from optimization (e.g., supply nets) by setting $c_e = 0$. We use the half-perimeter wire length (HPWL) metric to model the connectivity. For brevity, let $x_i^c=x_i + w_i/2, ~y_i^c=y_i+h_i/2$ be coordinates of the centroids of the rectangles. The final connectivity criterion element is formed as follows:
\begin{equation}
    \mathcal{L}_{conn} = \sum_{\forall e\in E} c_e \cdot (\max_{i \in L_e} x_i^c - \min_{i \in L_e} x_i^c + \max_{i \in L_e} y_i^c - \min_{i \in L_e} y_i^c)
\end{equation} 

\begin{figure}
        \centering
        \begin{subfigure}[b]{0.482\textwidth}
            \centering
            \includegraphics[width=\textwidth]{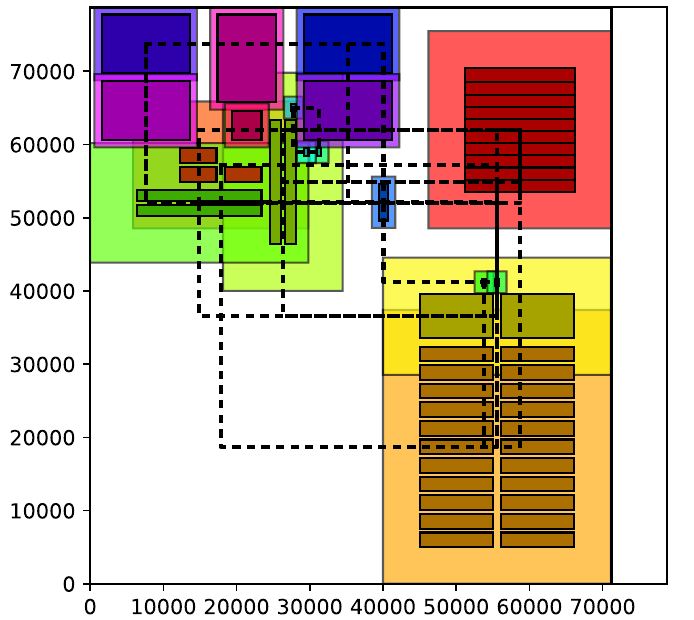}
            \label{fig:model_features_figure}
        \end{subfigure}
        \hfill
        \begin{subfigure}[b]{0.478\textwidth}  
            \centering 
            \includegraphics[width=\textwidth]{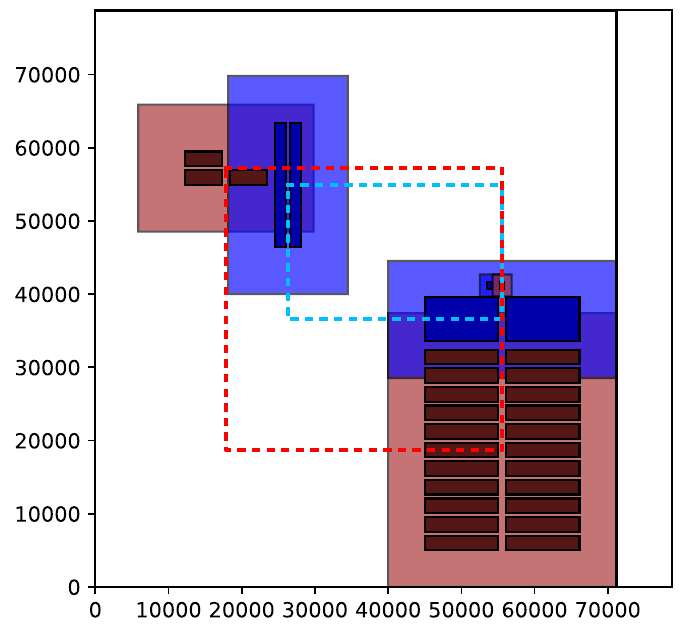}
            \label{fig:nets_figure}
        \end{subfigure}
        \caption{The left figure illustrates design features, with a large blockage area located in the bottom-left corner, and a symmetry group in the top-left corner. Nets are highlighted as dashed contours. Two nets and their rectangles are highlighted in the right figure.}
	\label{fig:model_ex}
\end{figure}

The definition of HPWL is simplified by placing the pins of the devices into their center, as in \citep{gaafteropt} - otherwise, offsets of the pins could be added into equations \eqref{hpwl:start}-\eqref{hpwl:end} to model the connectivity more precisely. See \cref{fig:model_ex} for an illustration of connectivity. To formulate this connectivity metric using ILP, we add four continuous variables $X^M_e, X^m_e, Y^M_e, Y^m_e \in \mathbb{R}$, per each net $e$. These variables represent the positions of the most extreme parts of the net - only the most distant elements contribute to the overall criterion. For example, $X^M_e$ holds the horizontal position of the centroid of the net's rectangle, which is located most to the right of the canvas. Then, we add the following constraints:
\begin{align}
    X^M_e \ge x_i + w_i/2 & & \forall i \in L_e ~\forall e \in E \label{hpwl:start}\\
    X^m_e \le x_i + w_i/2 & & \forall i \in L_e ~\forall e \in E \\
    Y^M_e \ge y_i + h_i/2 & & \forall i \in L_e ~\forall e \in E\\
    Y^m_e \le y_i + h_i/2 & & \forall i \in L_e ~\forall e \in E \label{hpwl:end}
\end{align}
\begin{equation}
    \mathcal{L}_{conn} = \sum_{\forall e \in E} c_e \cdot \left(X^M_e - X^m_e + Y^M_e - Y_e^m\right)
\end{equation}

We also calculate the normalization constant $S_{conn}$, which we use in the final criterion. We define it using the cost of each net $e$ as:
\begin{equation}
    S_{conn} = \sum_{\forall e \in E} c_e
\end{equation}

The mentioned features are demonstrated in \cref{fig:model_ex}, where the shown placement consists of 18 devices and topological structures with allowed pocket merging (see the bottom-right corner with yellow and orange rectangles overlapping). The blockage area occupies the bottom-left corner. The symmetry group with a vertical axis consisting of two symmetry pairs and two self-symmetric rectangles is shown in the top-left corner. Each net of the design is visualized as the dashed contour, which corresponds to the smallest bounding box containing all centroids of the net's rectangles. Two such nets are highlighted in \cref{fig:model_ex}.

\subsubsection{Optional criteria}
We may want to urge the solver to put specific pairs of rectangles closer or further from each other without explicitly defining the distance. We achieve this by introducing the proximity criterion. For each such concerned pair of rectangles, we define their associated cost $c_{prox}^{i,j}$, which can be either positive (to decrease their respective distance) or negative (to increase it). Then, we form the proximity criterion as follows:
\begin{equation}
    \mathcal{L}_{prox} = \sum_{\forall (i,j):c_{prox}^{i,j}\neq 0} c_{prox}^{i,j} \cdot d_{i,j}
\end{equation}
where $d_{i,j}$ is variable associated with the distance between rectangles $i$ and $j$. The distance can be formulated as Manhattan or Quasi-Euclidean, and between centroids or two closest points, as we have shown in our previous work \citep{icores23}. However, when used with a negative cost, we need to introduce additional binary variables. We demonstrate it with the Manhattan distance. For example, to constrain variable $d_{i,j}^x$ to contain the horizontal distance between the rectangles' centroids, the following constraints are needed, where $b_{i,j}^x \in \left\{0,1\right\}$.
\begin{align}
    x_i + w_i/2 - x_j - w_j/2& \le d_{i,j}^x \le & x_i + w_i/2 - x_j - w_j/2 + &M \cdot b_{i,j}^x&\\
    x_j + w_j/2 - x_i - w_i/2& \le d_{i,j}^x \le & x_j + w_j/2 - x_i - w_i/2 + &M \cdot (1 - b_{i,j}^x)&
\end{align}

This way, $d_{i,j}^x = |x_i + w_i/2 - x_j - w_j/2|$. With variables $d_{i,j}^x,d_{i,j}^y$, we obtain the Manhattan distance as $d_{i,j} = d_{i,j}^x + d_{i,j}^y$.

Since the optimized design is often just a single component of a larger IC, we need to consider its connectivity with other components. To do so, we form the interface connectivity criterion $\mathcal{L}_{inter}$. The designer can define the side of the placement where the interface net entry point is. The entry point is modeled as a dimensionless point attached to a specified side, but its exact position on the side is not fixed. We account for this with an additional continuous variable. Then, we optimize the distance between the entry point and connected rectangles to account for routing between different components of the IC. We define the distance in a point-to-point manner. Similar to the proximity criterion, each connection between the entry point and rectangle is associated with its cost, and the total criterion $\mathcal{L}_{inter}$ is obtained by summing weighted distances. 

Both the proximity criterion $\mathcal{L}_{prox}$ and the interface connectivity criterion $\mathcal{L}_{inter}$ are normalized with $S_{prox},S_{inter}$ respectively. We obtain these constants by summing the costs of the criteria elements. In the case of the $S_{prox}$, we sum the absolute values to account for the possible negative costs: $S_{prox} = \sum_{\forall (i,j)} |\,c_{prox}^{i,j}\,|$.

 The final criterion is obtained by combining all the criterion elements together with controllable costs $c_{\mathrm{cost}}$:
\begin{equation}\label{cost_eq}
    \mathcal{L} = c_{area}\cdot \mathcal{L}_{area} + \frac{c_{conn}}{S_{conn}}\cdot\mathcal{L}_{conn} + \frac{c_{prox}}{S_{prox}}\cdot\mathcal{L}_{prox} + \frac{c_{inter}}{S_{inter}}\cdot\mathcal{L}_{inter}
\end{equation} 

Normalization constants are used to make the tunable costs less sensitive to varying numbers of nets, proximity pairs, and interface connections. Further in the text, we do not consider the optional criteria $\mathcal{L}_{prox},~\mathcal{L}_{inter}$ during experiments. While they offer more control, we focus on the half perimeter and HPWL, which are common and important metrics in literature and industry. However, blockage areas and symmetry groups were considered in the experiments since they have a direct effect on placement being feasible rather than being an additional element of the criterion. Furthermore, the ILP solver used as a baseline is warm-started using the FDGD-based method described in \citep{icores23}. 

\section{Proposed Solution}
\label{sec:heur_const}
\subsection{Constructive heuristic}\label{sec:constr_descp}
\subsubsection{Outline of the heuristic}\label{sec:startheur}
The core of our solution is a constructive heuristic that transforms the input vector of numbers, denoted as a chromosome consisting of genes, to the individual solution (placement). This mapping is a crucial part of metaheuristics and local search. Our constructive heuristic is based on the mapping procedure of \citep{kubalik}. It uses an indirect representation of the placement. Each rectangle of the problem is associated with a triple of genes:

\begin{itemize}
	\item position - constructive heuristic places rectangles one by one; position genes are used to sort the rectangles for this procedure - the lower the value of the gene, the earlier the rectangle is placed
	\item variant - encodes a specific variant of the rectangle; the chosen variant is used within the particular solution, similarly to \citep{kubalik2}
	\item direction - determines whether we first start to search for a feasible point in the horizontal or vertical direction (see \cref{sec:sliding})
\end{itemize}

Therefore, we need $3\cdot n$ genes to represent $n$ rectangles. Furthermore, a single additional gene, which is used to perform priority modulation described later, can be appended to the chromosome as well; thus, the total length of the chromosome would be $3\cdot n + 1 $. Each gene is a real number from the interval between 0 and 1, as in \citep{ga_rcpsp}.

Now, we describe the proposed constructive procedure. Firstly, the initial point set $P$ is created, only containing the origin point (0,0). We first place objects with pre-defined coordinates (e.g., the blockage area), and we add their corners' coordinates to $P$ during this initialization phase. In each iteration of the main loop, the not-yet-placed rectangle with the lowest value of the position gene is selected, and the placement direction and the rectangle variant are determined from their respective genes. Then, we investigate each point of $P$ if the currently selected rectangle can be placed nearby (how we find a feasible position is described later). Unlike in \citep{kubalik}, where the first feasible point is chosen, we evaluate each feasible point and greedily place the rectangle in the position that increases the criterion value of the partial placement the least. 

\begin{figure}
	\centering
	\begin{subfigure}{.48\textwidth}
		\centering
		\includegraphics[width=.92\linewidth]{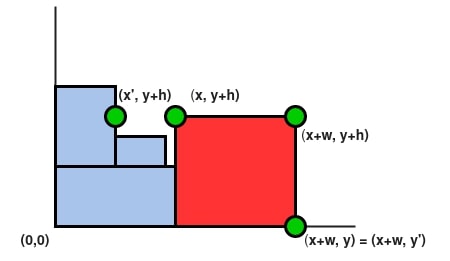}
		\caption{New points generated by placing the red rectangle}
		\label{ga_pointset}
	\end{subfigure}\hfill
	\begin{subfigure}{.4\textwidth}
		\centering
		\includegraphics[width=.9\linewidth]{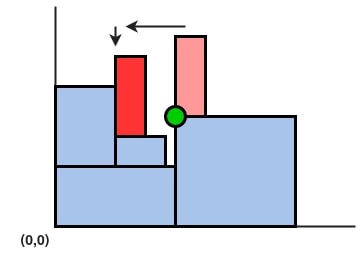}
		\caption{Sliding procedure performed on the red rectangle, starting at green point}
		\label{ga_placement}
	\end{subfigure}
	\caption{Visualization of the point generation and alternative coordinate calculation. Note that the calculation of the alternative coordinates is done for each point in $P$.}
	\label{ga_vis}
\end{figure}

After placing the rectangle, we generate new points to expand $P$. Whenever a new rectangle is placed at coordinates $(x,y)$, only up to 5 new points are added to point set $P$. Rectangle's corners (except the bottom-left one) are always added. We also add the intersection with the nearest obstacle while moving from points $(x+w,\,y)$, $(x,\, y+h)$ in a vertical, respectively, horizontal direction, with coordinates $(x+w,\, y')$, $(x',\, y+h)$. An example of new points added to $P$ is shown in \cref{ga_pointset}.

\subsubsection{Rectangle sliding}\label{sec:sliding}
Let $c$ be the rectangle that needs to be placed in the current iteration of the constructive heuristic. We investigate each point in $P$ and try to place the rectangle in its vicinity. Firstly, we compensate for the fact the points of $P$ are generated with an assumption that the minimum allowed distance between rectangles is 0. Whenever the minimum allowed distance $a_{c,r}$ between the current rectangle $c$ and the rectangle which generated the point $r$ is $a_{c,r}>0$, we move the point vertically or horizontally by $a_{i,j}$ so at least these two rectangles do not collide. If $a_{i,j} < 0$ (i.e., pocket merging), we do not modify the coordinates. Let $(x_p,y_p)$  be the point's updated coordinates after this step.

Then, a method based on bottom-left packing heuristic \citep{Hopper1999AGA} finds a near feasible point around the $(x_p,y_p)$. The method works in two phases; firstly, we modify the $x_p$ while fixing the value of $y_p$, and then vice versa. If the value of the rectangle's direction gene is greater than 0.5, we start with $y_p$ modification instead. We refer to this as rectangle sliding.

Now, we describe rectangle sliding with a fixed value of $y_p$. Firstly, we filter out already placed rectangles that could not collide with the current rectangle $c$ thanks to the coordinate $y_p$ alone. The rest of the placed rectangles are split into two sets, $R_L$ and $R_R$. Rectangle $r$ belongs to $R_L$ if $x_r + w_r/2 \leq x_p + w_c/2$; otherwise, it belongs to $R_R$. Then, we try to place the current rectangle so it has rectangles from $R_L$ to its left and those from $R_R$ to its right. This is done by computing:
\begin{align}
   x_L =& \max \left\{ \left\{ x_r + w_r +a_{c,r} ~|~ r \in R_L \right\} ~\cup~ \left\{ 0  \right\}  \right\} \\
   x_R =& \min \left\{ \left\{ x_r - w_c - a_{c,r} ~|~ r \in R_R \right\} ~\cup~ \left\{ \infty \right\} \right\}
\end{align}

If $x_L > x_R$, we recognize the situation as infeasible. Otherwise, we set $x_p = x_L$ and continue with the modification of $y_p$. If both phases succeed, we are left with feasible coordinates to place the rectangle $c$.

An example of the sliding procedure is shown in \cref{ga_placement}. The green starting point was moved first horizontally until the obstacle was reached and then vertically. However, always picking the bottom-left position could be detrimental to connectivity. Thus, the temporary point created after the first phase of the sliding procedure is also considered for placing the rectangle $c$.

\subsubsection{Priority modulation}
As described until this point, connectivity does not directly influence the heuristic. However, when the focus is put on minimizing connectivity, it is useful to modify the heuristic to account for it. We do this using priority modulation. The priority modulation factor $p_m$ can either be fixed beforehand or its value is set by the appended priority modulation gene, see \cref{sec:startheur}. After a rectangle is placed, we iterate through its nets and multiply each of its connected rectangles' position genes by $p_m$. Since $p_m \le 1$, the connected rectangles can be placed sooner. The intuition is that when the connected rectangles are processed one after another, there should be enough space to put them close to each other.

\subsubsection{Symmetry groups}
To handle the symmetry groups, we first place the symmetry group in a separate empty space. The same placement heuristic is used, with the modification that ensures that the axis of symmetry of the group is respected. Then, the symmetry group is returned to the main placement procedure as a single cohesive unit. Different symmetry group configurations can be explored by modifying associated position and variant genes of the group's rectangles. 

\subsubsection{Fitness of the individuals}
The individual's fitness is determined from the placement, following the definition of criterion \eqref{cost_eq} of \cref{sec:ilp_approach}. We use the half perimeter approximation of placement's area to compare our results directly with the ILP baseline of \cref{sec:ilp}. The procedure creates a placement that satisfies the constraints. An exception is the aspect ratio constraint, which is not directly enforced. Instead, when the aspect ratio constraint is violated, we multiply the final criterion value by $2.5$, which makes the search algorithms prefer placements that respect the constraint. 

\subsubsection{Complexity and visualization}
The proposed constructive heuristic has cubic complexity $O(n^3)$, given $n$ rectangles to be placed. This is comparable with the complexity of the Bottom Left Fill (BLF) constructive heuristic \citep{packing_heur_survey}. However, \citep{blf_quadratic} has shown quadratic time complexity implementation of BLF. Nevertheless, we could not exploit the more effective implementation due to the differences between our problem and plain rectangle packing.

\begin{figure}
        \centering
        \begin{subfigure}[b]{0.45\textwidth}
            \centering
            \includegraphics[width=\textwidth]{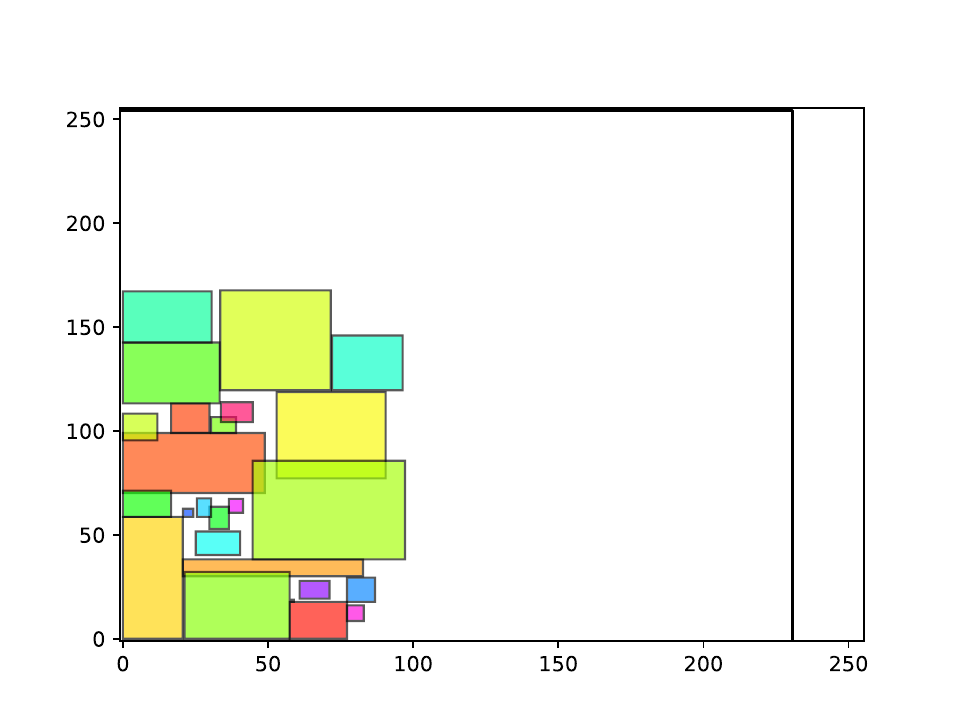}
            \caption{Partial placement with 25 \% of rectangles}    
            \label{fig:constr0}
        \end{subfigure}
        \hfill
        \begin{subfigure}[b]{0.45\textwidth}  
            \centering 
            \includegraphics[width=\textwidth]{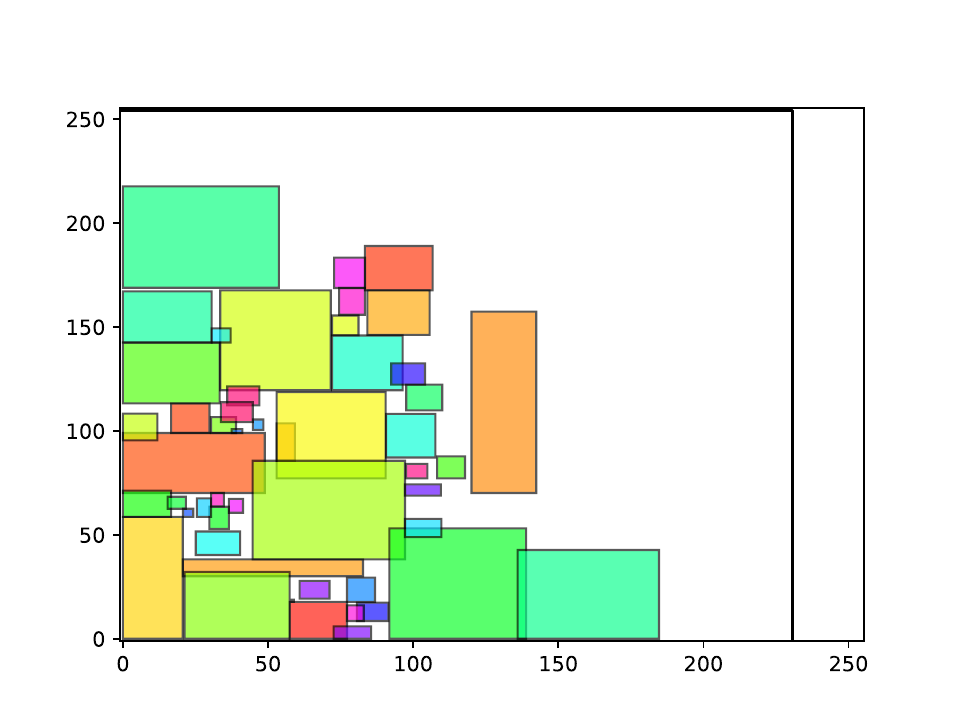}
            \caption{Partial placement with 50 \% of rectangles}   
            \label{fig:constr1}
        \end{subfigure}
        \vskip\baselineskip
        \begin{subfigure}[b]{0.45\textwidth}   
            \centering 
            \includegraphics[width=\textwidth]{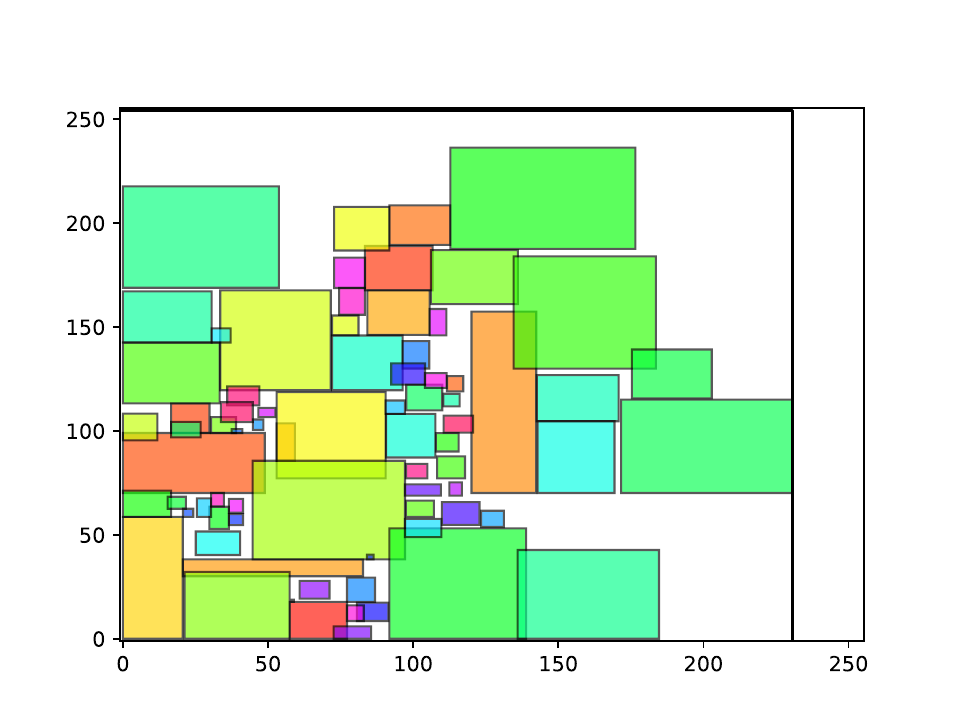}
            \caption{Partial placement with 75 \% of rectangles}   
            \label{fig:constr2}
        \end{subfigure}
        \hfill
        \begin{subfigure}[b]{0.45\textwidth}   
            \centering 
            \includegraphics[width=\textwidth]{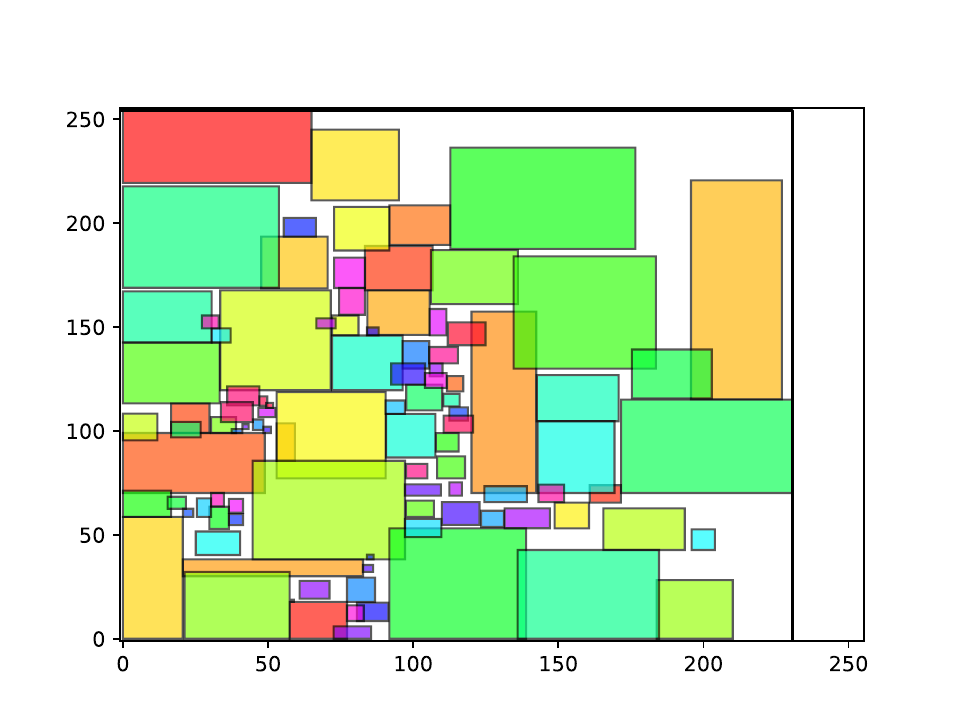}
            \caption{Completed placement}     
            \label{fig:constr3}
        \end{subfigure}
        \caption{Partial placements created during the run of the constructive heuristic.}
        \label{fig:constr}
\end{figure}

The heuristic is visualized in \cref{fig:constr}. Each figure corresponds to the partial placement, \cref{fig:constr0} contains 25 \% rectangles with the lowest value of the position gene, and \cref{fig:constr3} is the complete placement, containing 100 \% of rectangles. The construction starts from the origin (0,0) and iteratively fills the canvas.

\subsection{Metaheuristics}
\label{sec:meta_const}
Our proposed constructive heuristic maps chromosomes to feasible placements. However, a search must be performed to find a chromosome associated with a high-quality solution. 

\subsubsection{Genetic Algorithm}
\label{sec:ga_alg}
\begin{algorithm}
	\caption{Single generation of the genetic algorithm \citep{kubalik}}\label{alg:ga}
	\begin{algorithmic}
		\STATE list $L$ $\gets$ \{\}
		\REPEAT
  \STATE select parents $p_1$, $p_2$ from current population
		\IF {rand() $\le P_C$}
		\STATE child $c$ $\gets$ crossover of $p_1$ and $p_2$
		\IF {rand() $\le P_M$}
		\STATE $c$ $\gets$ mutation of $c$
		\ENDIF
		\ELSE
		\STATE child $c$ $\gets$ mutation of $p_1$
		\ENDIF
		\STATE push $c$ into $L$
        \UNTIL{size $s$ children are generated}
        \STATE list $E$ $\gets$ elite part of current population
        \RETURN best $s$ individuals from $L~\cup~E$
  
	\end{algorithmic}
\end{algorithm}

GA is a well-known population-based metaheuristic for solving black-box optimization problems. We utilized the GA's pipeline of \citep{kubalik}; the pseudocode of single generation is outlined in \cref{alg:ga}. We start with a randomly generated initial population, and part of it is optionally modified in a way similar to \citep{Hopper1999AGA}. Each rectangle's variant gene is set to select the most square-like variant, and their position genes are multiplied by a factor of $k_i / n$, where $k_i$ is the index of the rectangle $i$ if we sorted them in descending order of area. Thus, larger rectangles should be placed earlier than the smaller ones.

Then, we apply evolutionary operators in each generation. The selection procedure implements deterministic tournament selection with tournament size $T$. A two-point crossover is used. When mutated, the chromosome randomly modifies each of its genes with a probability of $0.1$. These operators are linked to a pair of algorithm's hyper-parameters $P_C,\, P_M \in \left[0;1\right]$. Finally, the elite part of the current population is propagated to the next generation directly.

In \cref{ga_3d}, we illustrate how the GA searches the solution space. Each point corresponds to the chromosome, projected to 2D space ($u,v$) by the Principal Component Analysis (PCA). The vertical axis corresponds to the criterion value of the associated placement. The color of the points corresponds to the generation in which the associated chromosome was created. We can see the criterion value of the points with the same color gets lower in the latter generations.

\subsubsection{Covariance Matrix Adaptation Evolution Strategy}
\label{sec:covmax}
CMA-ES \citep{hansen_cma} is a metaheuristic mostly used for continuous optimization problems. CMA-ES and its derivatives are considered state-of-the-art in the case of black-box optimization \citep{meta_comp}, especially when non-separable or ill-conditioned criterion functions are considered.

Starting with the initial individual, CMA-ES samples new individuals based on the mean vector and covariance matrix. The mean vector and covariance matrix are updated as the algorithm progresses, and the change is determined by the quality of the newly sampled individuals. The advantage from the user's perspective is the lack of tunable hyper-parameters of the CMA-ES. We only need to supply the starting individual and initial step size $\sigma$, which controls how far from the mean the new individuals are sampled; other parameters can be determined automatically.

\begin{figure}
	\centering
	\begin{subfigure}{.95\textwidth}
		\centering
		\includegraphics[width=.8\linewidth]{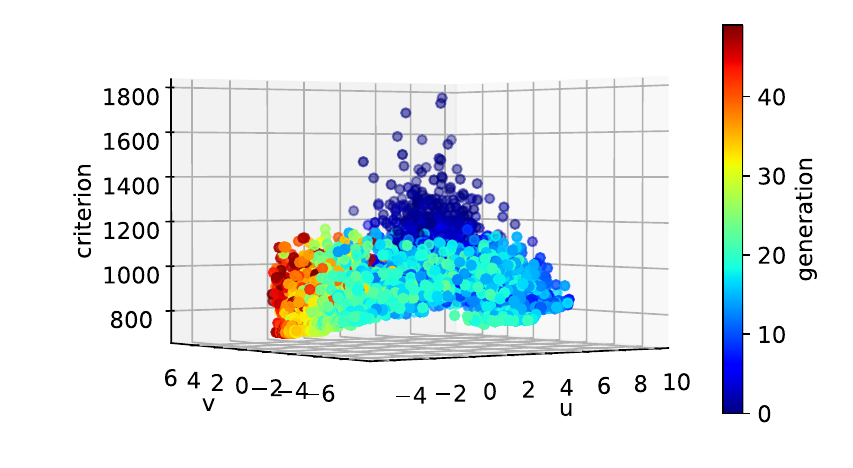}
		\caption{Genetic algorithm}
		\label{ga_3d}
	\end{subfigure}
    \vskip\baselineskip
	\begin{subfigure}{.95\textwidth}
		\centering
		\includegraphics[width=0.8\linewidth]{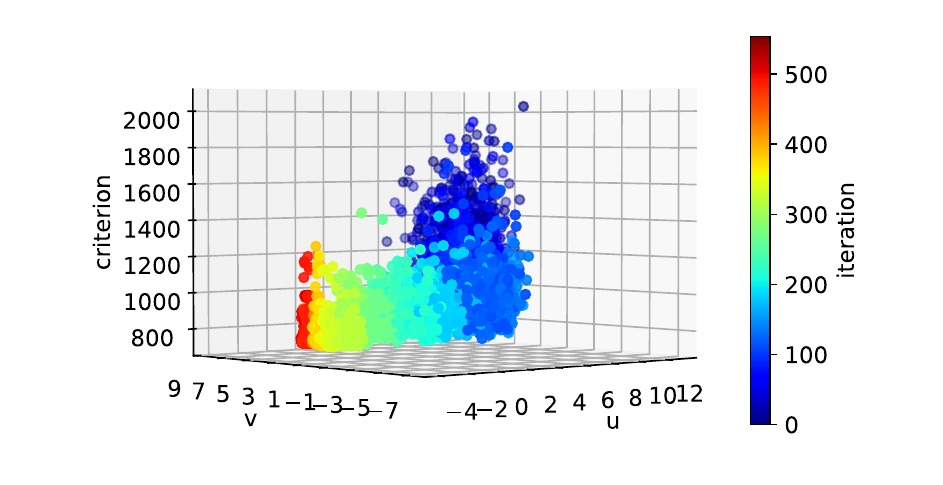}
        \caption{CMA-ES}
        \label{cma_3d}
	\end{subfigure}
	\caption{Chromosomes from each generation, projected using PCA to two dimensions $u,v$, with the criterion on vertical axis}
	\label{fig:ga_vis}
\end{figure}

The behavior of the CMA-ES is illustrated in \cref{cma_3d}. As we did with GA, the chromosomes obtained in each iteration were projected using PCA. The more compact distribution of individuals in the latter iterations demonstrates how the algorithm thoroughly searches the close neighborhood of the best-so-far solution - the covariance matrix was modified as the optimization ran.

\subsection{Local search and detailed optimization}
\label{sec:local_lp_text}
\begin{figure}
    \centering
    \includegraphics[width=0.99\textwidth]{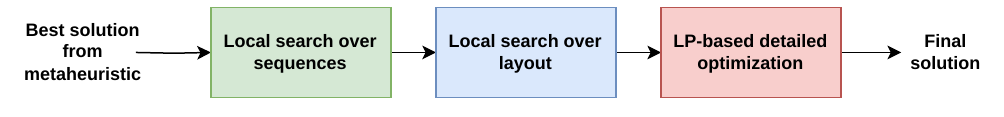}
    \caption{Improvement pipeline}
    \label{fig:impro-pipeline}
\end{figure}

We propose the following pipeline, visualized in \cref{fig:impro-pipeline}, to further optimize the solution. The first phase in the pipeline performs the local search over sequences (i.e., over chromosomes possibly missed by the metaheuristic). We modify the chromosome of the best solution, trying other variants of each rectangle one by one, and use the constructive heuristic to evaluate the new solution. For less complex instances, we perform a 2-opt-like position local search as well by trying to swap the values of position genes between pairs of rectangles.

\begin{figure}
        \centering
        \begin{subfigure}[b]{0.4\textwidth}
            \centering
            \includegraphics[width=\textwidth]{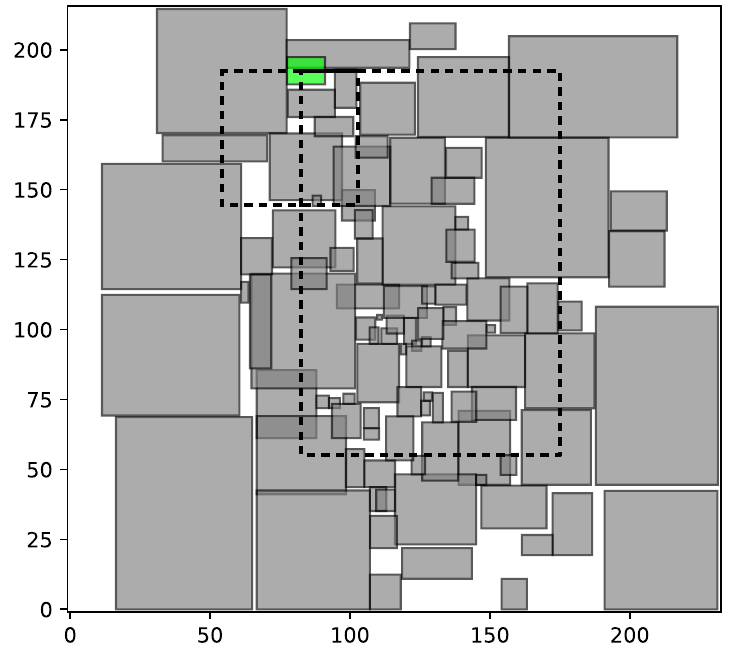}
            \caption{Placement before local search over layout, $\mathcal{L}_{conn} = 15135$.}    
            \label{fig:lslayout0}
        \end{subfigure}
        \hfill
        \begin{subfigure}[b]{0.4\textwidth}  
            \centering 
            \includegraphics[width=\textwidth]{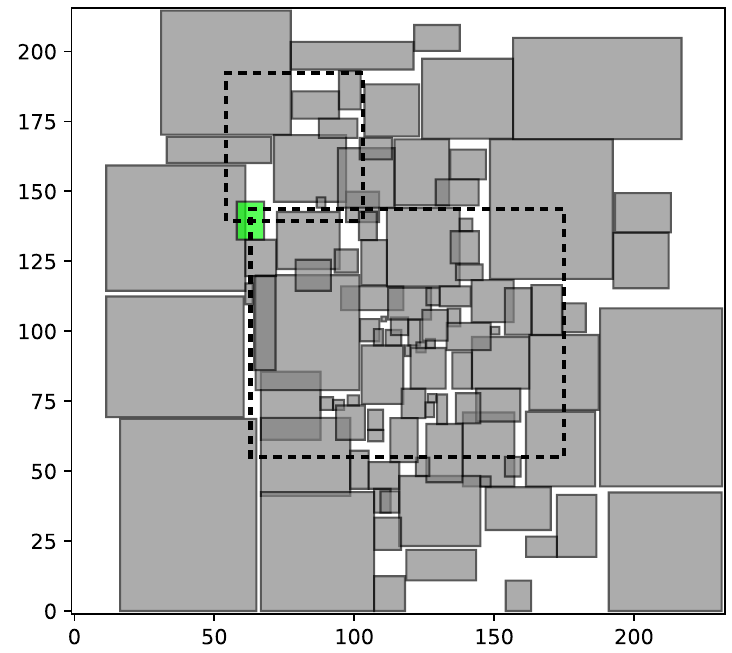}
            \caption{Placement after local search over layout, $\mathcal{L}_{conn} = 14752$.} 
            \label{fig:lslayout1}
        \end{subfigure}
        \caption{Illustration of the importance of the local search over the layout. The green rectangle was previously sub-optimally placed with respect to its nets (black dashed contours).} 
        \label{fig:lslayout}
\end{figure}

The second phase performs the local search directly over the layout as in \citep{grasp_packing}. We do this by fixing the placement and constructing a new point set $P$ given the corners of the placed rectangles. Then, for a selected rectangle, we use the rectangle sliding approach of \cref{sec:constr_descp} to try to find a new position that would lower the criterion. We do this iteratively for each rectangle (and its variants) and each point of $P$. This makes it possible for rectangles to be placed in positions otherwise inaccessible due to the iterative nature of the constructive heuristic. While the changes in the placement are subtle, they may reduce the HPWL significantly, as shown in \cref{fig:lslayout}.

Finally, Linear Programming (LP) detailed optimization based on \citep{gaafteropt} is performed. We modify the ILP formulation of \cref{sec:ilp} to not contain binary variables. To achieve this, we set the variant variables $s_i^k$ to only consider the variant currently used in the placement. Then we determine for each pair of rectangles which relative position constraint \eqref{pos1} - \eqref{rel2_eq} has the largest amount of slack, and we set $r_{i,j}^k=1$ for the corresponding relative position variable. With these actions, we are left with the LP model. The solver can only slightly change the coordinates of rectangles since their relative positions are fixed. We can solve it efficiently, potentially improving the quality of the solution.

\section{Reinforcement Learning Parameter Control}
\label{sec:rl_control_text}
Due to the recent successes of machine learning, we attempted to leverage it in our work in the following way. We used reinforcement learning, specifically Deep Q-Network (DQN), to control the hyper-parameters of the GA to yield, on average, better results \citep{param_control_overview}. 

This was investigated before in \citep{rl_basic}, where Q-learning was used to select search operators of GA. Since Q-learning handles only discrete input space (e.g., a fixed number of states in a card game), its generalization is needed to capture the continuous state representation (e.g., the continuous position of the car). DQN uses a neural network to approximate Q-values, the expected reward for choosing a specific action in a given state. Each available action is associated with a Q-value.

\subsection{Integrating DQN into the placement GA}
We chose to control the tournament ratio $t_T$ and crossover probability $P_C$ of the GA during evolution. Other parameters, such as elite size, could be considered as well. We periodically modified the hyper-parameters of GA with a period called environment-step. Whenever the number of generations equal to the environment-step passes, the agent performs an action to modify the hyper-parameters. Based on previous experiments, we set the environment-step to 4 to shorten the number of transitions between starting and terminal states; with a shorter period, we were not able to train the network properly.

Environment represents the GA population for a specific placement problem. Whenever the environment is restarted, a new placement problem and initial GA population are generated. The state representation was based on \citep{dqnga}, but only population features are considered. Twelve features in total are used, each normalized to the interval from 0 to 1. Features describe the size of the instance, chosen connectivity cost, the current generation's average criterion value and its standard deviation, the number of generations that have already evolved, the number of generations without the improvement, and statistics about the instance's rectangle sizes - percentage of multi-variant rectangles, and average and median size of the rectangle.

We set the reward returned by the environment to:
\begin{equation}
	r_t = \frac{\text{best}_{t-1} - \text{best}_t}{\text{best}_{\text{1}}}
\end{equation} where $\text{best}_t$ denotes the criterion value of the best individual in generation $t$. When an environment-step larger than one is considered, the reward is accumulated for each in-between generation. One episode of the environment refers to the whole run of the single problem instance until the final generation.

We have chosen the DQN reinforcement learning algorithm due to its simplicity and because it was recently used in the context of evolutionary algorithms as well (see \citep{dqnga}). Other methods, such as DDPG, TD3, or A2C, are much more complex, or need large amounts of iterations to train, or work in an online manner. However, DQN works with discrete actions only. Action space was therefore hand-picked, limited to 8 actions not to overwhelm the agent during exploration. We selected such action values to offer an agent a set of diverse options to select from:

\begin{itemize}
	\item tournament size ratio $t_T$ - $\lbrace 0.02,0.05,0.1,0.2\rbrace$
	\item crossover probability $P_C$ - $\lbrace 0.3,0.8\rbrace$
\end{itemize} 

Note that the actual tournament size $T$ used during selection is equal to:
\begin{equation}
    T=\lceil\text{population\_size}\cdot t_T\rceil
\end{equation}

A fully connected neural network with a single hidden layer was used. The input layer was connected to the hidden layer with 40 neurons, and the hidden layer was connected to the output layer, with each neuron corresponding to one of the actions. Non-linearity was introduced with ReLU. The output layer served as a Q-value estimator.

The training pipeline was based on \citep{chapman_lechner}. Experiments were carried out with 1000 episodes - thus, 1000 random synthetic examples were generated. The population size was set to 300, and the population evolved for a random number of generations, with the upper bound being 300. Each time 2500 new transition samples were obtained and added to the replay memory, the target network was updated. Discount factor $\gamma$ was set to 0.99, thus ensuring the rewards acquired in later generations still propagate to the earlier. The Adam optimizer (with learning rate $\alpha=0.001$) and Huber loss between current and estimated Q-values were used for training. $\epsilon$-greedy with linearly decaying $\epsilon$ approach was used for exploration.

\section{Experimental Evaluation}
\label{sec:experiments}
\subsection{Methodology}
This project was implemented in Python 3.8, with the computation-intensive constructive heuristic implemented in C. We utilized the Gurobi solver v9.5.1 \citep{gurobi} to solve the LP and ILP models. We used TensorFlow \citep{tensorflow2015-whitepaper} for machine learning and CMA-ES implementation of PyCMA library \citep{hansen2019pycma}. The experiments were performed on Intel Core i7-1255U. We used the ILP model warm-started by our FDGD approach \citep{icores23} as a baseline solution, utilized only for comparison.

\subsection{Overview of the synthetically generated instances}
\label{sec:synth_over}
To compare our developed approach and its modifications easily, we created several sets of synthetically generated instances inspired by real-life problems. Instance sets contain either 20 or 60 instances. Each instance contains a number of rectangles specific to its set, as shown in \cref{tab:synth_overview}; half of the rectangles were defined with multiple variants, and the other half only allowed rotation. Instances contained 0, 1, or 2 blockage areas. Each was randomly generated and blocked approximately 15 \% of rectangles. The minimum distance between each pair of rectangles was randomly generated, including the negative distances, to model the pocket merging - ensuring our problem's core feature is present among the instances. Afterward, nets were generated to model the connectivity between devices.

\begin{table}
	\centering
    \begin{adjustbox}{width=0.8\textwidth}
	\begin{tabular}{c | c c c c | c}
	dataset & \# instances & \# rectangles & symmetry & \# nets & \\
	\hline
	$S_{\textrm{50}}$ & 60 & 20, 30, 50 & No & 5 - 12 & \\
	$S_{\textrm{50}}^{\textrm{dense}}$ & 60 & 20, 30, 50 & No & 10 - 25 & \\
	$S_{\textrm{100}}$  & 20 & 100 & No & 25 & \\
	$S_{\textrm{double}}$  & 20 & 200 & No & 50 &   \begin{tabular}{@{}c@{}}2 copies of \\ $S_{\textrm{100}}$ instances \end{tabular} \\
	$S_{\textrm{tetra}}$  & 20 & 200 & No & 48 & \begin{tabular}{@{}c@{}}4 copies of \\ $S_{\textrm{50}}$ instances \end{tabular} \\
	$S_{\textrm{50}}^{\textrm{sym}}$  & 60 & 34 - 85 & Yes & 5 - 12 & \\
	$S_{\textrm{200}}^{\textrm{sym}}$ & 20 & 226 - 285 & Yes & 100 & \\
	\end{tabular}
    \end{adjustbox}
	\caption{Description of synthetically generated datasets}
	\label{tab:synth_overview}
\end{table}

Note that instances of sets $S_{\textrm{double}}$ and $S_{\textrm{tetra}}$ were generated by combining multiple copies of the identical smaller instance to create a larger one; connectivity within each copy was preserved, and no additional nets were introduced. Sets $S_{\textrm{50}}^{\textrm{sym}}$ and $S_{\textrm{200}}^{\textrm{sym}}$ were generated with several symmetry groups with 10+ rectangles. Finally, to compare the methods on more and less densely connected instances, two smaller sets $S_{\textrm{50}},~S_{\textrm{50}}^{\textrm{dense}}$ are used.

\subsection{Analysis and comparison of our approaches}
\label{sec:synth_exps}
In the rest of this section, we experimented with synthetically generated instances to study our proposed solution. In the following experiments, the criterion costs were set to $c_{area} = 1$ (which remained the same across all experiments), and $c_{conn} = 8$. We chose these values to primarily study the scenario when the connectivity is prioritized.

\subsubsection{Constructive heuristic parameters}
\label{sec:param_choice}
Firstly, we determined the constructive heuristic parameters by a limited computational study. Our point evaluation, described in \cref{sec:constr_descp}, is crucial; for instances of $S_{\textrm{50}}$, the criterion was improved by 10 \% when compared with the original method of \citep{kubalik}.

\begin{figure}
        \centering
        \begin{subfigure}[b]{0.4\textwidth}
            \centering
            \includegraphics[width=\textwidth]{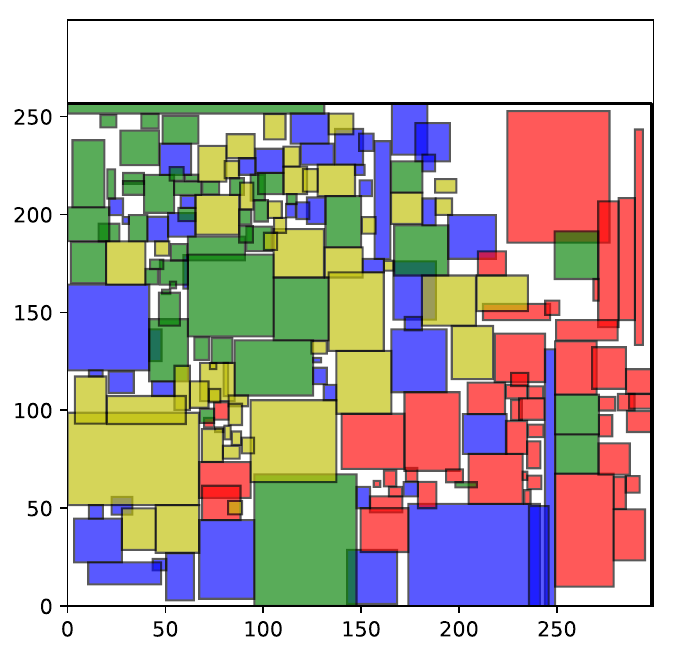}
            \caption{Placement without using the priority modulation. $\mathcal{L} = 1814$.}    
        \end{subfigure}
        \hfill
        \begin{subfigure}[b]{0.4\textwidth}  
            \centering 
            \includegraphics[width=\textwidth]{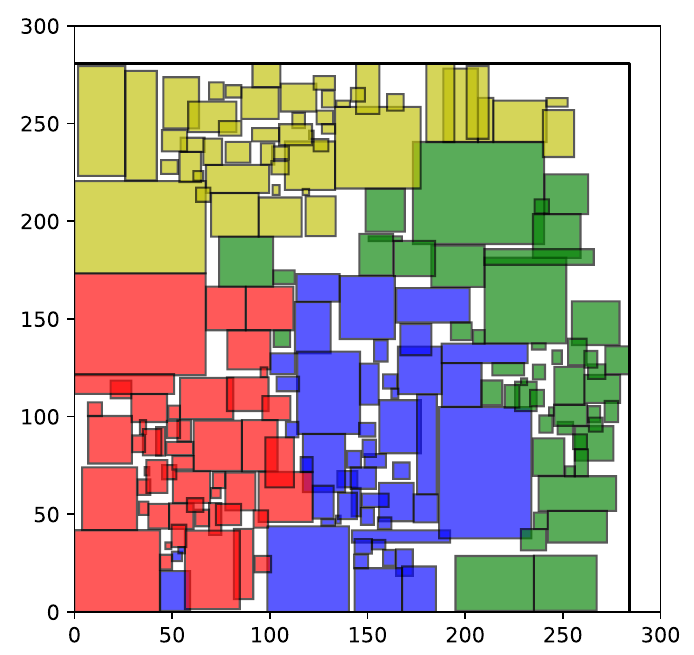}
            \caption{Placement using the priority modulation. $\mathcal{L} = 1311$.} 
        \end{subfigure}
        \caption{Placement obtained without and with priority modulation, on instance from set $S_{\textrm{tetra}}$. Only the same-colored rectangles are connected with nets.} 
        \label{fig:why-use-pmod}
\end{figure}

The priority modulation was necessary since large separable instances of $S_{\textrm{double}}$ and $S_{\textrm{tetra}}$ were otherwise hard to solve - it was difficult for metaheuristics to find the correct sequence of rectangles, that would bundle each separate group together in the final placement. When priority modulation with a sufficiently small coefficient was used, a much better value of the criterion was achieved. This is shown in \cref{fig:why-use-pmod}. There, nets are present only between the same-colored rectangles. We can see that priority modulation, given the same computation budget, found a much better solution and was able to put the connected rectangles close to each other.

We employed the priority modulation gene as a part of the chromosome to be adapted on the run. While we could achieve slightly better results by manually tuning its value depending on the instance, the self-adapted approach worked sufficiently well across all datasets. We ran experiments on all the datasets with $c_{conn} \in \left\{0.1,1,8\right\}$. For comparison, we tested GA with fixed priority modulation parameter $p_m\in \left\{0.1, 0.3, 0.5, 0.8, 1.0\right\}$. In the worst case, the self-adapted approach averaged criterion value 1 \% larger than the GA with the best value of $p_m$ for a given instance set and $c_{conn}$. However, no fixed value was consistently better; e.g., $p_m=0.3$ worked well for $S_{\textrm{tetra}}$, but not in other cases.  

\subsubsection{Effect of the local search and detailed optimization}
\label{sec:local_lp_eval}
\begin{table}
    \centering
    \begin{adjustbox}{width=0.7\textwidth}
    \begin{tabular}{c| c c c | c}
         & \multicolumn{3}{c|}{improvement per technique (\%)} &  \\
         dataset & seq LS & layout LS & LP opt & overall improvement (\%) \\
        \hline
$S_\textrm{50}$ & 1.00 & 0.87 & 0.78 & 2.65\\
$S_{\textrm{50}}^{\textrm{dense}}$ & 1.33 & 0.87 & 0.48 & 2.68\\
$S_\textrm{100}$& 3.26 & 1.46 & 0.24 & 4.96\\
$S_\textrm{double}$ & 2.42 & 2.33 & 0.20 & 4.95\\
$S_\textrm{tetra}$& 2.49 & 2.98 & 0.41 & 5.88\\
$S_\textrm{50}^{\textrm{sym}}$ & 2.64 & 0.66 & 0.44 & 3.74\\
$S_\textrm{200}^{\textrm{sym}}$ & 2.38 & 2.07 & 0.13 & 4.57\\
    \end{tabular}
    \end{adjustbox}
    \caption{Average effect of each improvement technique and average total improvement per dataset. Note that improvement of the individual technique is calculated with respect to its input solution, either the output of the GA or the output of the previous technique.}
    \label{tab:ls_impro_results}
\end{table}

As described in \cref{sec:local_lp_text}, we use three consecutive improvement techniques to improve solutions found by our proposed methods: local search over sequences, local search over layout, and LP-based detailed optimization.

We compared the quality of the initial solution found using the GA and the solution after we used the improvement techniques. In \cref{tab:ls_impro_results}, we present the percentage decrease of the criterion, averaged for each dataset. We also report improvement produced by each technique separately, relative to the GA's output criterion. We can see that all the methods improved the solution. The large effect of the search over sequence was expected; it runs first and requires more computation time. Overall, the improvement phase leads to the criterion being better by 2-5 \%. However, this is true only for tested costs $c_{area} = 1,\,c_{conn}=8$. For example, with $c_{conn}=0$, LP-based detailed optimization would have essentially no effect.

\subsubsection{Metaheuristics comparison}
\label{sec:metaheur}
We investigated the suitability of both discussed metaheuristics by evaluating them on datasets described in \cref{sec:synth_over}. Each experiment ran for a specific amount of time, as shown in the \cref{tab:meta_results}.

Hyper-parameters of the metaheuristics are shown in \cref{tab:meta_params}; the GA values correspond to those used in \citep{kubalik}, due to its success on a similar FLP problem; we verified that the algorithm produces acceptable result with several short experiments. The population size of GA varied depending on the dataset, with up to 250 generations to be evolved. If time was left, the population was reinitialized, and computation continued until the time limit was reached. Furthermore, 20 \% of the initial population was seeded as described in \cref{sec:ga_alg}. The population size of CMA-ES was selected automatically by the algorithm. 

\begin{table}
	\centering
    \begin{adjustbox}{width=0.7\textwidth}
	\begin{tabular}{c c}
	\multicolumn{2}{c}{GA}\\
\hline
tournament ratio & 0.02\\
elite ratio & 0.05 \\
$P_C$ & 0.8 \\
$P_M$ & 0.1\\
seed ratio & 0.2 \\
\end{tabular}
\quad
	\begin{tabular}{c c}
	\multicolumn{2}{c}{CMA-ES}\\
\hline
initial $\sigma$ & 0.25\\\\
initial solution & \begin{tabular}{@{}c@{}} best solution by GA\\ after 10 generations \end{tabular} \\
\end{tabular}
\end{adjustbox}
	\caption{Metaheuristics settings}
	\label{tab:meta_params}
\end{table}

Results are evaluated using the average relative difference (aRD) of the criterion, calculated for method $m$ and dataset $S$, defined as:
\begin{equation}
    \textrm{aRD}_S^m = \frac{1}{|S|}\cdot\sum_{i \in S}\frac{\mathcal{L}^{i, m} - \mathcal{L}^{i, {best}}}{\mathcal{L}^{i, {best}}} \cdot 100~[\%]
\end{equation}
where $\mathcal{L}^{i,m}$ is the value of criterion achieved on instance $i$ by method $m$, and $\mathcal{L}^{i,best}$ is the value of criterion of the best-known solution (among studied methods). Therefore, aRD refers to the ratio of the method's and best-known solution's criterion values averaged over the entire dataset. The best hits metric (\# best) tells us how many times a method achieved the best value of the criterion.

Results of this experimentation are shown in \cref{tab:meta_results}. On average, our heuristic algorithms outperformed the ILP baseline. ILP came closest in the case of $S_{\textrm{50}}^{\textrm{dense}}$ and $S_{\textrm{tetra}}$ datasets.

\begin{table}
	\centering
    \begin{adjustbox}{width=0.93\textwidth}
	\begin{tabular}{c c c | c c | c c | c c c}
		\multicolumn{3}{c|}{dataset} & \multicolumn{2}{c|}{ILP}& \multicolumn{2}{c|}{CMA-ES} & 	 	\multicolumn{3}{c}{GA} \\
		name & \# inst & time {[min]}  & aRD & \# best & aRD & \# best & pop size & aRD &\# best \\
	\hline
$S_{\textrm{50}}$ & 60 & 10 & 7.11 & 8 & \textbf{0.74} & \textbf{43} & 300 & 4.18 & 9\\
$S_{\textrm{50}}^{\textrm{dense}}$ & 60 & 10 & 2.82 & 17 & \textbf{0.51} & \textbf{39} & 300 & 2.80 & 4\\
$S_{\textrm{100}}$ & 20 &20 & 4.81 & 0 & \textbf{0.52} & \textbf{12} & 500 & 1.56 & 8\\
$S_{\textrm{double}}$ & 20 & 40 & 4.17 & 3 & 0.68 & \textbf{12} & 500 & \textbf{0.51} & 10\\
$S_{\textrm{tetra}}$ & 20 &40 & 3.35 & 6 & 1.11 & 5 & 500 & \textbf{0.78} & \textbf{10} \\
$S_{\textrm{50}}^{\textrm{sym}}$ & 60& 10 & 28.10 & 0 & \textbf{0.08} & \textbf{56} & 300 & 4.15 & 4\\
$S_{\textrm{200}}^{\textrm{sym}}$ & 20& 40 &35.44 & 0 & 0.47 & 9 & 500 & \textbf{0.22} & \textbf{16} \\
	\end{tabular}
 \end{adjustbox}
	\caption{Averaged results obtained on synthetically generated instances, comparing metaheuristics and baseline ILP approach.}
	\label{tab:meta_results}
\end{table}

When we compare the results of GA and CMA-ES, we see they are competitive. On smaller instances with a lower number of rectangles, it seems CMA-ES outperforms the GA. However, this changes for larger instances, and GA seems to outperform CMA-ES by a slight margin. It seems that the CMA-ES is able to search the neighborhood of the current solution quite well, as long as the complexity of the problem is not too large. After a certain point, the random moves introduced by the genetic algorithm seem to help the search more than the exhaustive sampling of local solutions.

\subsubsection{Effect of the parameter control}
\label{sec:rl_control_eval}
We present our two best-performing parameter control models, described in \cref{sec:rl_control_text}. The first model, whose results are denoted as DQN-GA1, used all features outlined in \cref{sec:rl_control_text}, while DQN-GA2 had its instance size feature set to 0 for each training problem instance to study its effect on the overall results. We evaluated both models on each dataset, using the same metrics as in \cref{sec:metaheur}. Averaged results are shown in \cref{tab:dqn_results}.

\begin{table}
	\centering
    \begin{adjustbox}{width=1\textwidth}
	\begin{tabular}{c c c | c c| c c | c c | c c}
		\multicolumn{3}{c|}{dataset} & \multicolumn{2}{c|}{Our GA} & 	\multicolumn{2}{c|}{DQN-GA1} & 	\multicolumn{2}{c|}{DQN-GA2} & 	\multicolumn{2}{c}{GA with $P_C=0.3$}\\
		name & \# inst & time {[min]} & aRD & \# best & aRD & \# best & aRD & \# best& aRD & \# best\\
	\hline
 $S_{\textrm{50}}$ & 60 & 10 & 2.13 & 9 & 1.66 & 17 & 1.40 & \textbf{21} & \textbf{1.32} & 13  \\
$S_{\textrm{50}}^{\textrm{dense}}$ & 60&10& 2.05 & 11 & 1.60 & 14 & \textbf{1.51} & \textbf{20} & 1.55 & 15\\
	$S_{\textrm{100}}$  &  20 & 20& 2.23 & 5 & 2.14 & 5 & 2.44 & \textbf{6} & \textbf{2.12} & 4\\
	$S_{\textrm{double}}$  & 20 & 40&\textbf{1.01} & \textbf{6} & 1.69 & 4 & 1.43 & \textbf{6} & 1.26 & 5 \\
	$S_{\textrm{tetra}}$  & 20&40 & 1.29 & \textbf{6} & 1.25 & 5 & \textbf{1.21} & 3 & 1.26 & \textbf{6}\\
	$S_{\textrm{50}}^{\textrm{sym}}$ & 60&10  & 2.39 & 6 & 2.14 & 15 & 1.58 & \textbf{20} & \textbf{1.24} & 19 \\
	$S_{\textrm{200}}^{\textrm{sym}}$ & 20 &40& 1.70 & 5 & 1.34& 4 & 1.96 & 1 & \textbf{0.56} & \textbf{10}\\
	\end{tabular}
 \end{adjustbox}
	\caption{Averaged results obtained on synthetically generated instances, comparing baseline GA and DQN-controlled GAs.}
	\label{tab:dqn_results}
\end{table}

\begin{figure}
\centering
\begin{subfigure}{.5\textwidth}
  \centering
  \includegraphics[width=.98\linewidth]{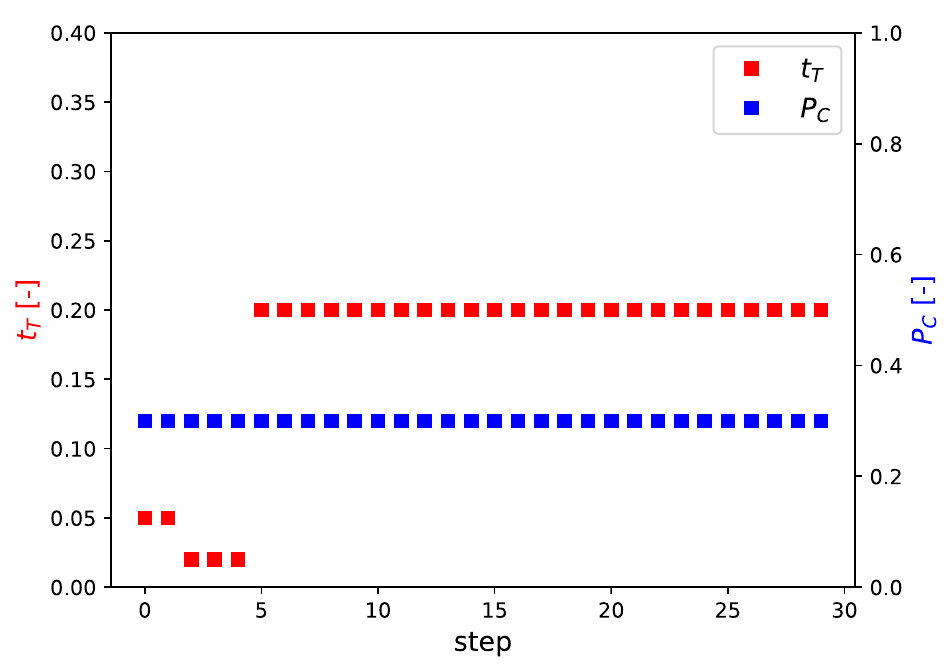}
  \caption{DQN-GA1}
  \label{fig:dqncourse1}
\end{subfigure}\hfill
\begin{subfigure}{.5\textwidth}
  \centering
  \includegraphics[width=.98\linewidth]{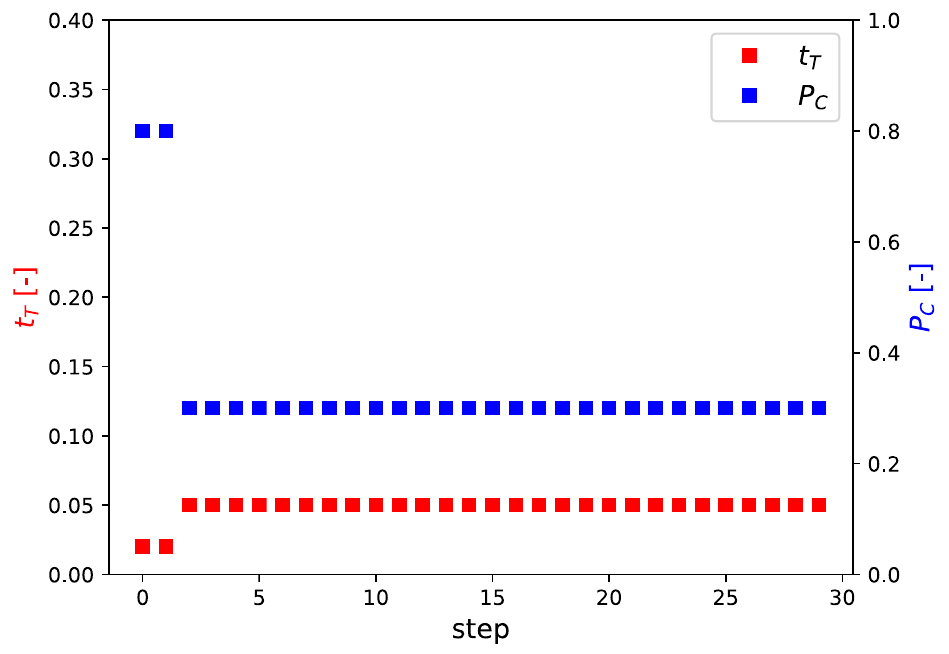}
  \caption{DQN-GA2}
  \label{fig:dqncourse2}
\end{subfigure}
\caption{Values of GA hyper-parameters during evolution. The values in each step were selected by the DQN agent.}
\label{fig:dqncourse}
\end{figure}

On average, DQN-GA1 outperformed the original GA with fixed parameters, even though large instances were under-represented in training. DQN-GA2 outperformed both our fixed-parameter GA and DQN-GA1 on several sets, but especially on $S_{\textrm{200}}^{\textrm{sym}}$, it behaved rather worse. An example of the control pattern of both reinforcement learning models is shown in \cref{fig:dqncourse}. While the patterns were obtained on a single instance, they are typical for their respective DQN models across instances with varying sizes. 

We can observe that both models mostly selected action  $t_T=0.05$, $P_C=0.3$, which corresponds more to parameter selection rather than control; the agent was not able to learn a dynamic policy given a training environment we used. For comparison, we also ran an experiment with GA with fixed parameters, but using $P_C=0.3$ instead, whose results are reported in the last column of \cref{tab:dqn_results}; the results are in accordance with the RL experiments and show we could even more slightly improve the results by using the discovered values of parameters. 

To summarise, the slight ($< 1$\%) improvements provided by the proposed reinforcement learning approach do not justify the resources required for the offline training of the agent. Online adaptive methods, such as work of \citep{adaptivegenetic,adaptivegenetic2}, could be used without the training overhead. However, thanks to this experiment, we found out we could slightly improve our method's overall results if we select $P_C=0.3$, which can be observed by comparing the first and last columns of \cref{tab:dqn_results}.

\subsection{MCNC and GSRC}
\begin{table}
	\centering
    \begin{adjustbox}{width=1\textwidth}
	\begin{tabular}{c |c  c|c c c|c c c|c c c|  c c c} 
	    &\multicolumn{2}{c|}{\citep{camose_mcnc}} & \multicolumn{3}{c|}{\citep{symmetry_island_mcnc}} & \multicolumn{3}{c|}{\citep{hiearchichal_mcnc}} &  \multicolumn{3}{c|}{CMA-ES $c_{conn}=0$} & \multicolumn{3}{c}{CMA-ES $c_{conn}=2$} \\
	    &area & time &area& HPWL & time &area& HPWL & time&area & HPWL & time &area & HPWL & time \\ \hline
	    apte & \textbf{46.92} & 2 & 47.90 & - & 3 & 47.08 & 297.12 & 6 & \textbf{46.92} & 615.66 & 2 & 48.44 & \textbf{205.28} & 2\\
	    hp & \textbf{9.35} & 3 & 10.10 &-&16&9.57& \textbf{74.38} &32 & \textbf{9.35} & 170.27 & 2 & 9.46 & 82.36 & 2\\
	    ami33 & \textbf{1.21} & 19 &1.29&47.23&39&1.26&45.05&348 & 1.23& 91.63 &30 & 1.25 & \textbf{40.39} & 30 \\
	    ami49& 38.17 & 44&41.32 &769.99&96& 39.52 &763.93&559 & \textbf{37.64} & 1269.71 & 120& 40.49& \textbf{718.32} &120  \\
	\end{tabular}
    \end{adjustbox}
	\caption{MCNC benchmark instances results, area in mm$^2$, HPWL in mm, time in~s. \citep{camose_mcnc} utilized Intel i7 2.3 GHz, \citep{symmetry_island_mcnc} utilized Pentium4 3.2GHz, \citep{hiearchichal_mcnc} utilized Intel E5-2690 2.9GHz.}
	\label{tab:mcnc_comparison}
\end{table}

\begin{table}
	\centering
    \begin{adjustbox}{width=1\textwidth}
	\begin{tabular}{c |c c | c c c | c c c|c c c|  c c c} &\multicolumn{2}{c|}{\citep{gsrc_area_only}}
	    &\multicolumn{3}{c|}{\citep{gsrc_dp}} &\multicolumn{3}{c|}{\citep{gsrc_memetic}} &\multicolumn{3}{c|}{GA $c_{conn}=2$}  &\multicolumn{3}{c}{GA $c_{conn}=2$}\\
     &area & time & area & HPWL & time & area & HPWL & time & area & HPWL & time & area & HPWL & time \\\hline
     n100 & 0.194 & 69 & \textbf{0.180} & \textbf{101.51} & 48 & 0.201 & 165.30& 51 & 0.195 & 155.78 & 180 & 0.197 & 121.66 & 180 \\
     n200 & 0.195 & 162 & \textbf{0.183} & 289.72 & 1103 & 0.193 & 374.60& 167 & 0.190 & 330.06 & 197 & 0.191 & \textbf{262.89} & 197\\
     n300 & - & - & \textbf{0.276 }& 448.56 & 1743 & 0.310 & 436.40 & 464 & 0.296 & 559.02 & 291 & 0.299 & \textbf{416.10} & 291\\
	\end{tabular}
    \end{adjustbox}
	\caption{GSRC benchmark instances results, area in mm$^2$, HPWL in mm, time in~s. \citep{gsrc_area_only} used 2GHz Intel system, \citep{gsrc_dp} used 2.10 GHz AMD Ryzen5, and \citep{gsrc_memetic} used A1018 2.10 GHz.}
	\label{tab:gsrc_benchmark}
\end{table}

The MCNC benchmark set \citep{mcnc_bench} consists of 4 problem instances, namely \textbf{apte} (9 rectangles, 4 symmetric pairs), \textbf{hp} (11 rectangles, 4 symmetric pairs), \textbf{ami33} (33 rectangles, 3 symmetric pairs) and \textbf{ami49} (49 rectangles, 2 symmetric pairs). From GSRC benchmark we used instances \textbf{n100}, \textbf{n200}, \textbf{n300}, with 100, 200, 300 rectangles respectively. In the case of MCNC instances, we used CMA-ES, and for GSRC instances, due to their larger size, GA was used. Since there are no non-zero minimum distances between devices, features of our more general approach are not fully utilized. Therefore, we aim to show our solution's competitiveness even given a more specific problem statement. We performed experiments with two values of $c_{conn}$.

On the MCNC benchmark, we compare ourselves with the following methods: absolute representation approach of \citep{camose_mcnc}, topological representation approach of \citep{symmetry_island_mcnc}, and ILP-based approach of \citep{hiearchichal_mcnc}. On the GSRC benchmark, we compare ourselves with the approach of \citep{gsrc_area_only} producing sliceable floorplans, and B*-tree-based methods \citep{gsrc_dp,gsrc_memetic}.

\begin{figure}
\centering
\begin{subfigure}{.5\textwidth}
  \centering
  \includegraphics[width=.98\linewidth]{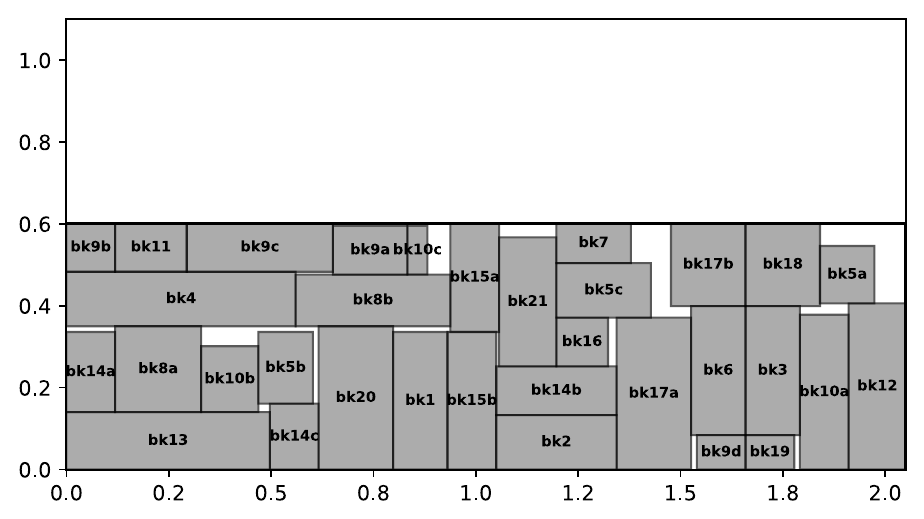}
  \caption{Solution for $c_{conn} = 0$}
  \label{fig:ami33a}
\end{subfigure}\hfill
\begin{subfigure}{.5\textwidth}
  \centering
  \includegraphics[width=.98\linewidth]{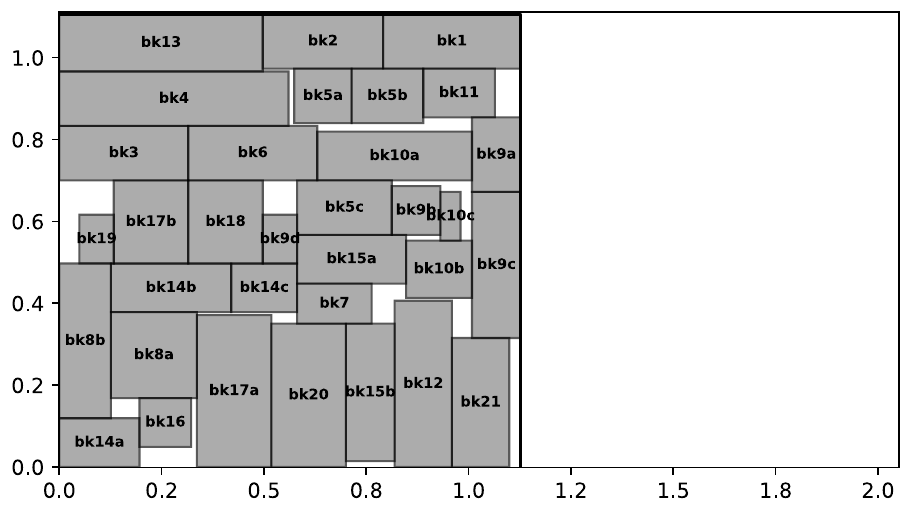}
  \caption{Solution for $c_{conn} = 2$}
  \label{fig:ami33b}
\end{subfigure}
\caption{Solutions found by CMA-ES for instance \textbf{ami33}.}
\label{fig:ami33}
\end{figure}

\begin{figure}
\centering
\begin{subfigure}{.5\textwidth}
  \centering
  \includegraphics[width=.7\linewidth]{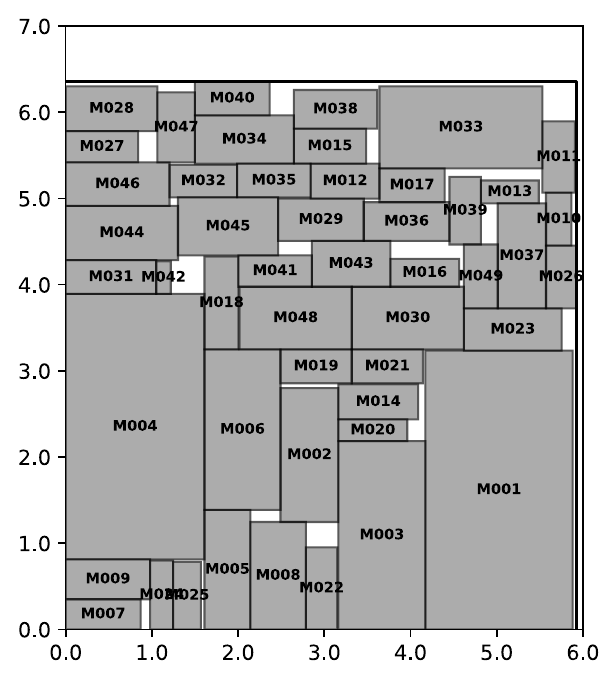}
  \caption{Solution for $c_{conn} = 0$}
  \label{fig:ami49a}
\end{subfigure}\hfill
\begin{subfigure}{.5\textwidth}
  \centering
  \includegraphics[width=.7\linewidth]{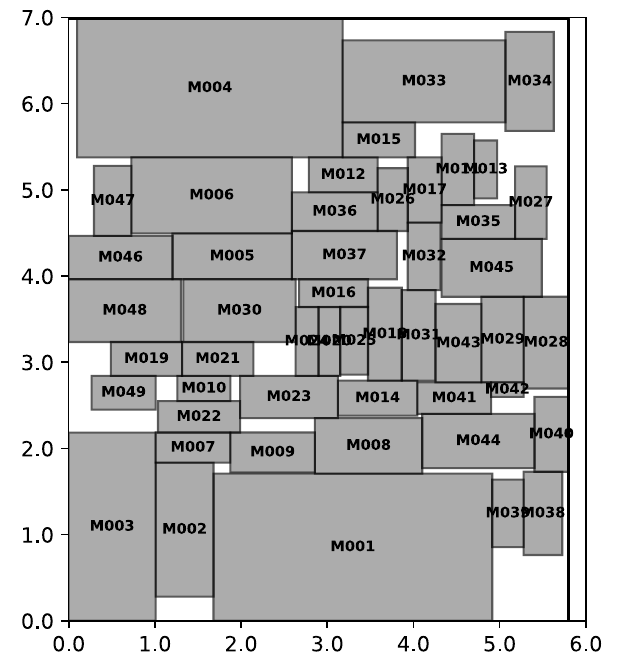}
  \caption{Solution for $c_{conn} = 2$}
  \label{fig:ami49b}
\end{subfigure}
\caption{Solutions found by CMA-ES for instance \textbf{ami49}.}
\label{fig:ami49}
\end{figure}

Approaches \citep{camose_mcnc,gsrc_area_only} considered only the placement area; therefore, the experiment with connectivity cost $c_{conn}=0$ is suitable for comparison. On MCNC instances, our method either outperformed or tied the results of \citep{camose_mcnc}, except for \textbf{ami33}. HPWL optimization was included in \citep{symmetry_island_mcnc,hiearchichal_mcnc}, so we investigated whether we were able to find overall better or at least non-dominated solutions. We found a non-dominated solution in each case and an overall better solution for instance \textbf{ami33}. Solutions for instances \textbf{ami33} and \textbf{ami49} are shown in \cref{fig:ami33,fig:ami49}.

On GSRC instances, we outperformed the results of \citep{gsrc_area_only,gsrc_memetic}. However, in the case of \citep{gsrc_dp}, authors reported results with an area smaller than we were able to obtain, and in the case of the \textbf{n100} with a smaller value of HPWL as well. When we compare our results on the two larger instances, we can see that we were eventually able to shorten the connectivity. Considering the fact that our algorithm was devised to handle the non-uniform minimum distances, which was not applied to these examples, we were still able to find competitive solutions (at least in the wire-length sense).

\subsection{ALIGN placer and open-source instances}
ALIGN \citep{align} is a contemporary open-source layout system. The system contains circuit annotation, placement, and routing tools. We used the ALIGN annotation tool to discover symmetry groups and topological structures in the input instances to make the comparison between the solvers as fair as possible.

Due to the differences in the scopes of ALIGN and this paper, we evaluate our method on a set of ALIGN instances. These were two Operational Transconductance Amplifiers (OTA), a Double Tail Sense Amplifier (DTS-A), a Switched Capacitor Filter (SCF), and a Linear Equalizer (LE), each accessible from the ALIGN repository. The instances contained symmetry groups and topological structures. We manually retrieved minimum distances between devices from the reference solutions and sanitized the input data; this was done due to the differences in problem formulations and data format and made the results rather illustrative. Together with ALIGN and baseline ILP solutions, the best results produced by CMA-ES are shown in \cref{tab:align_table}.

\begin{table}
	\centering
 \small
 \begin{adjustbox}{width=0.9\linewidth}
	\begin{tabular}{c | c c c | c c c| c c c} 
		& \multicolumn{3}{c|}{ALIGN \citep{align}}& \multicolumn{3}{c|}{ILP}&\multicolumn{3}{c}{CMA-ES}\\ 
		instance & area& HPWL& time & area& HPWL & time & area & HPWL & time \\ \hline
 CC-OTA & 73.2 & \textbf{132.2}& 46.1 &\textbf{58.3} &141.4 & 6.0 & 59.4 & 142.1&6.8\\
 T-OTA & \textbf{16.9} & 28.7 & 19.0 & 18.6 & \textbf{28.5}& 0.3 & 18.7 & 29.8 & 1.0 \\
 DTS-A & 52.8 & \textbf{69.4}& 15.5 & \textbf{44.7} & 90.0 & 0.5&50.8&101.3&3.5 \\
 SCF & 1995.6 & \textbf{478.4}& 126.1 & 1963.4 & 485.7& 13.8 & \textbf{1808.9} & 479.3 & 10.9\\
 LE & 58.2 & \textbf{47.0}& 42.7 & \textbf{56.5} & 56.2 & 6.7 & 57.5 & 55.4 & 5.5\\
 \end{tabular}
 \end{adjustbox}
		\caption{Comparison of our approach with ALIGN \citep{align} and baseline ILP. Area in $\mu\textrm{m}^2$, HPWL in $\mu\textrm{m}$, time in s. Results obtained on Intel Core i7-1255U.}	
  \label{tab:align_table}
\end{table}

CMA-ES could approximately match the quality of both ALIGN and baseline ILP solutions; we view this as a success, given the differences in the problem scopes and the small size of the problems. Furthermore, CMA-ES also used the same amount of computation time as the ILP solver. The ALIGN placer ran until it terminated (to avoid possible unfair comparison due to the differences in solver settings and the underlying system), and its computation time was longer than ours in all cases. Finally, we outperformed our ILP baseline solution on instance SCF, which contained the largest number of devices, demonstrating that our approach is suitable for larger problems.

\subsection{Real-life benchmarks}
Finally, we evaluated our solution against the baseline ILP and manual placements on a batch of real-life problem instances. In total, 17 instances were obtained from STMicroelectronics company. Since the real-life instances contained up to 60 independent rectangles, we again selected CMA-ES as a metaheuristic. We ensured that the same computation time was given to CMA-ES and baseline ILP, up to 8 minutes per problem instance. However, several instances were so simple the Gurobi solver actually found the optimal solution. 

Three experiments were performed for each instance, with $c_{area}=1$ and $c_{conn} \in \lbrace 0.1,\, 1,\, 8\rbrace$. Aggregated results are shown in \cref{tab:cma_vs_ilp_vs_bipop}. In the table, data for both scenarios, with and without pocket merging, are present, thus increasing the number of tested instances from 17 to 34. Most of the real-life benchmarks contained only up to 40 rectangles. Thus, we split the result between instances with less than 20, and more than 20 rectangles. Exceptionally good performance of ILP solver for smaller instances is expected, as it even found optimal solutions in several cases. 
On larger instances, the value of aRD is consistently smaller for the CMA-ES, and it was also able to find the same or more overall better solutions than the ILP.

\begin{table}
\centering
    \begin{adjustbox}{width=0.7\textwidth}
\begin{tabular}{c|c c| c c | c c | c c}
& \multicolumn{4}{c|}{ $<$ 20 rectangles} & \multicolumn{4}{c}{20+ rectangles}\\
 & \multicolumn{2}{c|}{ILP} & \multicolumn{2}{c|}{CMA-ES}& \multicolumn{2}{c|}{ILP} & \multicolumn{2}{c}{CMA-ES} \\
    $c_{conn}$ & aRD & \# best & aRD & \# best  & aRD & \# best & aRD & \# best \\
	\hline
 0.1 & \textbf{0.16} & \textbf{13} & 0.88 & 3 & 2.83 & 5 & \textbf{0.35} & \textbf{13}\\
1 & \textbf{0.02}& \textbf{15} & 1.70 & 1 & 2.39 & \textbf{9} & \textbf{0.76} & \textbf{9}\\
8 & \textbf{0.15} & \textbf{15} & 8.58 & 1 & 2.96 & \textbf{10} & \textbf{1.93} & 8\\
\end{tabular}
\end{adjustbox}
\caption{Values of aRD and (\# best) metrics for baseline ILP and CMA-ES on real-life benchmarks. Results are divided based on the number of rectangles of the instances.}
\label{tab:cma_vs_ilp_vs_bipop}
\end{table}

\begin{table} 
\centering
    \begin{adjustbox}{width=1\textwidth}
\begin{tabular}{c|c c c|c c c|c c c|c c c}
	 & \multicolumn{3}{c|}{manual} & \multicolumn{9}{c}{CMA-ES} \\
	& \multicolumn{3}{c|}{}& \multicolumn{3}{c|}{$c_{conn} = 0.1$} & \multicolumn{3}{c|}{$c_{conn} = 1$} & \multicolumn{3}{c}{$c_{conn} = 8$}  \\
	instance &  W+H & area & HPWL &  W+H & area & HPWL & W+H & area & HPWL &   W+H & area & HPWL\\
	\hline
1& 158& 6118& 1850& 156& 6023& 2153& \textbf{157}& \textbf{6117}& \textbf{1843}& 172& 7340& 1630\\
2& 116& 2710& 1784& \textbf{88}& \textbf{1906}& \textbf{1385}& \textbf{93}& \textbf{2143}& \textbf{935}& 117& 3043& 852\\
3& 106& 2650& 906& \textbf{87}& \textbf{1818}& \textbf{868}& \textbf{87}& \textbf{1807}& \textbf{770}& \textbf{99}& \textbf{2326}& \textbf{619}\\
4& 129& 4096& 812& \textbf{112}& \textbf{3117}& \textbf{782}& \textbf{112}& \textbf{3117}& \textbf{782}& \textbf{124}& \textbf{3762}& \textbf{725}\\
5& 207& 8972& 13797& \textbf{159}& \textbf{6329}& \textbf{9377}& \textbf{159}& \textbf{6352}& \textbf{9199}& \textbf{175}& \textbf{7606}& \textbf{8166}\\
6& 178& 7698& 4039& \textbf{161}& \textbf{6460}& \textbf{3904}& \textbf{165}& \textbf{6643}& \textbf{3869}& \textbf{166}& \textbf{6828}& \textbf{3744}\\
7& 168& 6580& 2908& 162& 6512& 3607& 165& 6776& 2830& 172& 7391& 2441\\
8& 173& 7294& 1501& 160& 6371& 1658& \textbf{162}& \textbf{6402}& \textbf{1233}& 181& 8176& 1143\\
9& 243& 14129& 4705& 223& 12414& 6114& 226& 12731& 4720& 250& 15541& 4263\\
10& 205& 10214& 28386& 194& 9362& 40209& 197& 9703& 37539& 207& 10759& 30611\\
11& 225& 9922& 28527& 200& 9994& 22737& \textbf{208}& \textbf{9432}& \textbf{17437}& \textbf{215}& \textbf{9821}& \textbf{17003}\\
12& 155& 5953& 3824& \textbf{123}& \textbf{3803}& \textbf{2668}& \textbf{128}& \textbf{4100}& \textbf{2172}& \textbf{134}& \textbf{4443}& \textbf{2031}\\
13& 162& 6511& 2061& 153& 5851& 2537& 153& 5869& 2103& 174& 7561& 1854\\
14& 247& 15235& 2399& \textbf{185}& \textbf{8534}& \textbf{2249}& \textbf{190}& \textbf{8989}& \textbf{1563}& \textbf{211}& \textbf{10931}& \textbf{1352}\\
15& 123& 3758& 1619& 114& 3233& 1923& 114& 3250& 1803& 116& 3385& 1679\\
16& 232& 12397& 2676& 214& 11405& 2906& \textbf{223}& \textbf{12373}& \textbf{2384}& 231& 13315& 2046\\
17& 247& 12525& 4586& 218& 11792& 4817& \textbf{220}& \textbf{12015}& \textbf{3750}& 239& 14201& 3404\\
\hline
avg ratio 
& 1.00 & 1.00 & 1.00 & 0.88 & 0.83 & 1.03 & 0.90 & 0.85 & 0.87 & 0.97 & 1.00 & 0.77 \\
\end{tabular}
\end{adjustbox}
\caption{Values of $W+H$ in $\mu$m, area in $\mu\textrm{m}^2$, HPWL in $\mu$m, and an average ratio of design generated using CMA-ES metaheuristic and manual designs for real-life instances. Highlighted triples correspond to solutions dominating manual ones.}
\label{tab:realressize_bipop}
\end{table}

We also compare CMA-ES results with the manual designs in \cref{tab:realressize_bipop}. For each solution, the half perimeter of the bounding box, area, and HPWL metrics are reported, with results averaged and related to the manual designs in the last row. Furthermore, for 12 out of 17 instances, our approach found a solution dominating the manual design in all studied metrics. The industry experts from STMicroelectronics verified the results.

We found that the half-perimeter approximation of the area enabled us to successfully solve many instances. One notable exception is the 17th instance for $c_{conn}=8$, where the half-perimeter is significantly lower, but the area is larger. This was caused by the fact that the manual design had an aspect ratio of 0.4, being in the region where the approximation is no longer accurate enough.

\section{Conclusion}
\label{sec:conclusion}
In this paper, we modeled the placement process of the AMS ICs, trying to automate it for BCD technology, which includes complex proximity constraints, pockets, and their merging, as well as more standard constraints, like symmetry groups and blockage areas. We proposed a constructive placement heuristic to handle the mentioned constraints, and we used GA and CMA-ES metaheuristics to find high-quality solutions. We also described the reinforcement learning method for control of the GA's parameters during optimization.

We evaluated our methods on synthetically generated instances, well-known benchmarks, and real-life instances, considering both the HPWL and the placement area. From results obtained on synthetically generated instances, we determined how well the improvement techniques and the parameter control work. We also compared both metaheuristics with our baseline ILP solution.

We were competitive with successful contemporary approaches on MCNC and GSRC datasets. We also compared our methods with the contemporary open-source ALIGN placer to show the competitiveness for slightly different problem formulations. Finally, we evaluated our algorithm using a set of real-life instances with associated manual designs. For 12 out of 17 instances, our approach found a solution that outperformed the manual one regarding both the final area and HPWL connectivity. Finally, our industrial partner, STMicroelectronics, will be able to use our approach as a fast prototyping tool in production to reduce the cumbersome manual work.

\section*{Acknowledgments}
We want to thank STMicroelectronics, namely Dalibor Barri and Patrik Vacula, for their help regarding the problem formulation and for providing us with real-life industrial instances.

This work was supported by the Grant Agency of the Czech Republic under the Project GACR 22-31670S. This work was co-funded by the European Union under the project ROBOPROX (reg. no. CZ.02.01.01/00/22\_008/0004590). 

\bibliographystyle{elsarticle-num-names} 
\bibliography{refs}

\begin{thebibliography}{51}
\expandafter\ifx\csname natexlab\endcsname\relax\def\natexlab#1{#1}\fi
\providecommand{\url}[1]{\texttt{#1}}
\providecommand{\href}[2]{#2}
\providecommand{\path}[1]{#1}
\providecommand{\DOIprefix}{doi:}
\providecommand{\ArXivprefix}{arXiv:}
\providecommand{\URLprefix}{URL: }
\providecommand{\Pubmedprefix}{pmid:}
\providecommand{\doi}[1]{\href{http://dx.doi.org/#1}{\path{#1}}}
\providecommand{\Pubmed}[1]{\href{pmid:#1}{\path{#1}}}
\providecommand{\bibinfo}[2]{#2}
\ifx\xfnm\relax \def\xfnm[#1]{\unskip,\space#1}\fi
\bibitem[{Scheible and Lienig(2015)}]{survey}
\bibinfo{author}{J.~Scheible}, \bibinfo{author}{J.~Lienig},
\newblock \bibinfo{title}{Automation of analog {IC} layout: Challenges and solutions},
\newblock in: \bibinfo{booktitle}{Proceedings of the 2015 Symposium on International Symposium on Physical Design}, ISPD '15, \bibinfo{publisher}{Association for Computing Machinery}, \bibinfo{address}{New York, NY, USA}, \bibinfo{year}{2015}, p. \bibinfo{pages}{33–40}. \URLprefix \url{https://doi.org/10.1145/2717764.2717781}. \DOIprefix\doi{10.1145/2717764.2717781}.
\bibitem[{Munien and Ezugwu(2021)}]{survey_1dbpp}
\bibinfo{author}{C.~Munien}, \bibinfo{author}{A.~E. Ezugwu},
\newblock \bibinfo{title}{Metaheuristic algorithms for one-dimensional bin-packing problems: A survey of recent advances and applications},
\newblock \bibinfo{journal}{Journal of Intelligent Systems} \bibinfo{volume}{30} (\bibinfo{year}{2021}) \bibinfo{pages}{636--663}. \URLprefix \url{https://doi.org/10.1515/jisys-2020-0117}. \DOIprefix\doi{doi:10.1515/jisys-2020-0117}.
\bibitem[{Murata et~al.(1996)Murata, Fujiyoshi, Nakatake, and Kajitani}]{murata_sequence_pairs}
\bibinfo{author}{H.~Murata}, \bibinfo{author}{K.~Fujiyoshi}, \bibinfo{author}{S.~Nakatake}, \bibinfo{author}{Y.~Kajitani},
\newblock \bibinfo{title}{{VLSI} module placement based on rectangle-packing by the sequence-pair},
\newblock \bibinfo{journal}{IEEE Trans. Comput. Aided Des. Integr. Circuits Syst.} \bibinfo{volume}{15} (\bibinfo{year}{1996}) \bibinfo{pages}{1518--1524}.
\bibitem[{Balasa and Lampaert(1999)}]{sequence_pairs}
\bibinfo{author}{F.~Balasa}, \bibinfo{author}{K.~Lampaert},
\newblock \bibinfo{title}{Module placement for analog layout using the sequence-pair representation},
\newblock in: \bibinfo{booktitle}{Proceedings 1999 Design Automation Conference (Cat. No. 99CH36361)}, \bibinfo{year}{1999}, pp. \bibinfo{pages}{274--279}. \DOIprefix\doi{10.1109/DAC.1999.781325}.
\bibitem[{Ma et~al.(2011)Ma, Xiao, Tam, and Young}]{handling}
\bibinfo{author}{Q.~Ma}, \bibinfo{author}{L.~Xiao}, \bibinfo{author}{Y.-C. Tam}, \bibinfo{author}{E.~F.~Y. Young},
\newblock \bibinfo{title}{Simultaneous handling of symmetry, common centroid, and general placement constraints},
\newblock \bibinfo{journal}{IEEE Transactions on Computer-Aided Design of Integrated Circuits and Systems} \bibinfo{volume}{30} (\bibinfo{year}{2011}) \bibinfo{pages}{85--95}. \DOIprefix\doi{10.1109/TCAD.2010.2064490}.
\bibitem[{Dhar et~al.(2021)Dhar, Kunal, Li, Madhusudan, Poojary, Sharma, Xu, Burns, Harjani, Hu, Kirkpatrick, Mukherjee, Yaldiz, and Sapatnekar}]{align}
\bibinfo{author}{T.~Dhar}, \bibinfo{author}{K.~Kunal}, \bibinfo{author}{Y.~Li}, \bibinfo{author}{M.~Madhusudan}, \bibinfo{author}{J.~Poojary}, \bibinfo{author}{A.~K. Sharma}, \bibinfo{author}{W.~Xu}, \bibinfo{author}{S.~M. Burns}, \bibinfo{author}{R.~Harjani}, \bibinfo{author}{J.~Hu}, \bibinfo{author}{D.~A. Kirkpatrick}, \bibinfo{author}{P.~Mukherjee}, \bibinfo{author}{S.~Yaldiz}, \bibinfo{author}{S.~S. Sapatnekar},
\newblock \bibinfo{title}{{ALIGN}: A system for automating analog layout},
\newblock \bibinfo{journal}{IEEE Design \& Test} \bibinfo{volume}{38} (\bibinfo{year}{2021}) \bibinfo{pages}{8--18}. \DOIprefix\doi{10.1109/MDAT.2020.3042177}.
\bibitem[{Pisinger(2007)}]{pisinger_seqforstrip}
\bibinfo{author}{D.~Pisinger},
\newblock \bibinfo{title}{Denser packings obtained in o(n log log n) time},
\newblock \bibinfo{journal}{INFORMS Journal on Computing} \bibinfo{volume}{19} (\bibinfo{year}{2007}) \bibinfo{pages}{395--405}. \URLprefix \url{https://doi.org/10.1287/ijoc.1060.0192}. \DOIprefix\doi{10.1287/ijoc.1060.0192}. \href{http://arxiv.org/abs/https://doi.org/10.1287/ijoc.1060.0192}{{\tt arXiv:https://doi.org/10.1287/ijoc.1060.0192}}.
\bibitem[{Lin et~al.(2009)Lin, Chang, and Lin}]{symmetry_island_mcnc}
\bibinfo{author}{P.-H. Lin}, \bibinfo{author}{Y.-W. Chang}, \bibinfo{author}{S.-C. Lin},
\newblock \bibinfo{title}{Analog placement based on symmetry-island formulation},
\newblock \bibinfo{journal}{IEEE Transactions on Computer-Aided Design of Integrated Circuits and Systems} \bibinfo{volume}{28} (\bibinfo{year}{2009}) \bibinfo{pages}{791--804}. \DOIprefix\doi{10.1109/TCAD.2009.2017433}.
\bibitem[{S and MS(2023)}]{gsrc_dp}
\bibinfo{author}{D.~C. S}, \bibinfo{author}{S.~MS},
\newblock \bibinfo{title}{Multi-objective floorplanning optimization engaging dynamic programming for system on chip},
\newblock \bibinfo{journal}{Microelectronics Journal} \bibinfo{volume}{140} (\bibinfo{year}{2023}) \bibinfo{pages}{105942}. \URLprefix \url{https://www.sciencedirect.com/science/article/pii/S0026269223002550}. \DOIprefix\doi{https://doi.org/10.1016/j.mejo.2023.105942}.
\bibitem[{Shanthi et~al.(2022)Shanthi, Rani, and Rajaram}]{gsrc_memetic}
\bibinfo{author}{J.~Shanthi}, \bibinfo{author}{D.~G.~N. Rani}, \bibinfo{author}{S.~Rajaram},
\newblock \bibinfo{title}{An enhanced memetic algorithm using skb tree representation for fixed-outline and temperature driven non-slicing floorplanning},
\newblock \bibinfo{journal}{Integration} \bibinfo{volume}{86} (\bibinfo{year}{2022}) \bibinfo{pages}{84--97}. \URLprefix \url{https://www.sciencedirect.com/science/article/pii/S0167926022000384}. \DOIprefix\doi{https://doi.org/10.1016/j.vlsi.2022.04.001}.
\bibitem[{Strasser et~al.(2008)Strasser, Eick, Grab, Schlichtmann, and Johannes}]{representation_as_trees}
\bibinfo{author}{M.~Strasser}, \bibinfo{author}{M.~Eick}, \bibinfo{author}{H.~Grab}, \bibinfo{author}{U.~Schlichtmann}, \bibinfo{author}{F.~M. Johannes},
\newblock \bibinfo{title}{Deterministic analog circuit placement using hierarchically bounded enumeration and enhanced shape functions},
\newblock in: \bibinfo{booktitle}{2008 IEEE/ACM International Conference on Computer-Aided Design}, \bibinfo{year}{2008}, pp. \bibinfo{pages}{306--313}. \DOIprefix\doi{10.1109/ICCAD.2008.4681591}.
\bibitem[{Cohn et~al.(1991)Cohn, Garrod, Rutenbar, and Carley}]{koan_anagram}
\bibinfo{author}{J.~Cohn}, \bibinfo{author}{D.~Garrod}, \bibinfo{author}{R.~Rutenbar}, \bibinfo{author}{L.~Carley},
\newblock \bibinfo{title}{{KOAN/ANAGRAM II}: new tools for device-level analog placement and routing},
\newblock \bibinfo{journal}{IEEE Journal of Solid-State Circuits} \bibinfo{volume}{26} (\bibinfo{year}{1991}) \bibinfo{pages}{330--342}. \DOIprefix\doi{10.1109/4.75012}.
\bibitem[{Martins et~al.(2015)Martins, Lourenço, and Horta}]{camose_mcnc}
\bibinfo{author}{R.~Martins}, \bibinfo{author}{N.~Lourenço}, \bibinfo{author}{N.~Horta},
\newblock \bibinfo{title}{Multi-objective optimization of analog integrated circuit placement hierarchy in absolute coordinates},
\newblock \bibinfo{journal}{Expert Systems with Applications} \bibinfo{volume}{42} (\bibinfo{year}{2015}) \bibinfo{pages}{9137--9151}. \URLprefix \url{https://www.sciencedirect.com/science/article/pii/S0957417415005655}. \DOIprefix\doi{https://doi.org/10.1016/j.eswa.2015.08.020}.
\bibitem[{Xu et~al.(2019)Xu, Zhu, Liu, Lin, Li, Tang, Sun, and Pan}]{magical}
\bibinfo{author}{B.~Xu}, \bibinfo{author}{K.~Zhu}, \bibinfo{author}{M.~Liu}, \bibinfo{author}{Y.~Lin}, \bibinfo{author}{S.~Li}, \bibinfo{author}{X.~Tang}, \bibinfo{author}{N.~Sun}, \bibinfo{author}{D.~Z. Pan},
\newblock \bibinfo{title}{{MAGICAL}: Toward fully automated analog ic layout leveraging human and machine intelligence: Invited paper},
\newblock in: \bibinfo{booktitle}{2019 IEEE/ACM International Conference on Computer-Aided Design (ICCAD)}, \bibinfo{year}{2019}, pp. \bibinfo{pages}{1--8}. \DOIprefix\doi{10.1109/ICCAD45719.2019.8942060}.
\bibitem[{Lin et~al.(2022)Lin, Li, Fang, Madhusudan, Sapatnekar, Harjani, and Hu}]{date_placer_fdgd}
\bibinfo{author}{Y.~Lin}, \bibinfo{author}{Y.~Li}, \bibinfo{author}{D.~Fang}, \bibinfo{author}{M.~Madhusudan}, \bibinfo{author}{S.~S. Sapatnekar}, \bibinfo{author}{R.~Harjani}, \bibinfo{author}{J.~Hu},
\newblock \bibinfo{title}{Are analytical techniques worthwhile for analog {IC} placement?},
\newblock in: \bibinfo{booktitle}{2022 Design, Automation \& Test in Europe Conference \& Exhibition ({DATE})}, \bibinfo{year}{2022}, pp. \bibinfo{pages}{154--159}. \DOIprefix\doi{10.23919/DATE54114.2022.9774498}.
\bibitem[{Chen et~al.(2008)Chen, Jiang, Hsu, Chen, and Chang}]{ntuplace}
\bibinfo{author}{T.-C. Chen}, \bibinfo{author}{Z.-W. Jiang}, \bibinfo{author}{T.-C. Hsu}, \bibinfo{author}{H.-C. Chen}, \bibinfo{author}{Y.-W. Chang},
\newblock \bibinfo{title}{{NTUplace3}: An analytical placer for large-scale mixed-size designs with preplaced blocks and density constraints},
\newblock \bibinfo{journal}{IEEE Transactions on Computer-Aided Design of Integrated Circuits and Systems} \bibinfo{volume}{27} (\bibinfo{year}{2008}) \bibinfo{pages}{1228--1240}. \DOIprefix\doi{10.1109/TCAD.2008.923063}.
\bibitem[{Xu et~al.(2019)Xu, Li, Pui, Liu, Shen, Lin, Sun, and Pan}]{gaafteropt}
\bibinfo{author}{B.~Xu}, \bibinfo{author}{S.~Li}, \bibinfo{author}{C.-W. Pui}, \bibinfo{author}{D.~Liu}, \bibinfo{author}{L.~Shen}, \bibinfo{author}{Y.~Lin}, \bibinfo{author}{N.~Sun}, \bibinfo{author}{D.~Z. Pan},
\newblock \bibinfo{title}{Device layer-aware analytical placement for analog circuits},
\newblock in: \bibinfo{booktitle}{Proceedings of the 2019 International Symposium on Physical Design}, ISPD '19, \bibinfo{publisher}{Association for Computing Machinery}, \bibinfo{address}{New York, NY, USA}, \bibinfo{year}{2019}, p. \bibinfo{pages}{19–26}. \URLprefix \url{https://doi.org/10.1145/3299902.3309751}. \DOIprefix\doi{10.1145/3299902.3309751}.
\bibitem[{Xu et~al.(2017)Xu, Li, Xu, Sun, and Pan}]{hiearchichal_mcnc}
\bibinfo{author}{B.~Xu}, \bibinfo{author}{S.~Li}, \bibinfo{author}{X.~Xu}, \bibinfo{author}{N.~Sun}, \bibinfo{author}{D.~Z. Pan},
\newblock \bibinfo{title}{Hierarchical and analytical placement techniques for high-performance analog circuits},
\newblock in: \bibinfo{booktitle}{Proceedings of the 2017 ACM on International Symposium on Physical Design}, ISPD '17, \bibinfo{publisher}{Association for Computing Machinery}, \bibinfo{address}{New York, NY, USA}, \bibinfo{year}{2017}, p. \bibinfo{pages}{55–62}. \URLprefix \url{https://doi.org/10.1145/3036669.3036678}. \DOIprefix\doi{10.1145/3036669.3036678}.
\bibitem[{Marolt et~al.(2016)Marolt, Scheible, Jerke, and Marolt}]{swarm}
\bibinfo{author}{D.~Marolt}, \bibinfo{author}{J.~Scheible}, \bibinfo{author}{G.~Jerke}, \bibinfo{author}{V.~Marolt},
\newblock \bibinfo{title}{Swarm: A self-organization approach for layout automation in analog ic design},
\newblock \bibinfo{journal}{International Journal of Electronics and Electrical Engineering}  (\bibinfo{year}{2016}) \bibinfo{pages}{374--385}. \DOIprefix\doi{10.18178/ijeee.4.5.374-385}.
\bibitem[{Mirhoseini et~al.(2021)Mirhoseini, Goldie, Yazgan, Jiang, Songhori, Wang, Lee, Johnson, Pathak, Nazi, Pak, Tong, Srinivasa, Hang, Tuncer, Le, Laudon, Ho, Carpenter, and Dean}]{graph_placement}
\bibinfo{author}{A.~Mirhoseini}, \bibinfo{author}{A.~Goldie}, \bibinfo{author}{M.~Yazgan}, \bibinfo{author}{J.~W. Jiang}, \bibinfo{author}{E.~Songhori}, \bibinfo{author}{S.~Wang}, \bibinfo{author}{Y.-J. Lee}, \bibinfo{author}{E.~Johnson}, \bibinfo{author}{O.~Pathak}, \bibinfo{author}{A.~Nazi}, \bibinfo{author}{J.~Pak}, \bibinfo{author}{A.~Tong}, \bibinfo{author}{K.~Srinivasa}, \bibinfo{author}{W.~Hang}, \bibinfo{author}{E.~Tuncer}, \bibinfo{author}{Q.~V. Le}, \bibinfo{author}{J.~Laudon}, \bibinfo{author}{R.~Ho}, \bibinfo{author}{R.~Carpenter}, \bibinfo{author}{J.~Dean},
\newblock \bibinfo{title}{A graph placement methodology for fast chip design},
\newblock \bibinfo{journal}{Nature} \bibinfo{volume}{594} (\bibinfo{year}{2021}) \bibinfo{pages}{207--212}. \URLprefix \url{https://doi.org/10.1038/s41586-021-03544-w}. \DOIprefix\doi{10.1038/s41586-021-03544-w}.
\bibitem[{Berger et~al.(2009)Berger, Schr{\"o}der, and K{\"u}fer}]{CSP_ortho}
\bibinfo{author}{M.~Berger}, \bibinfo{author}{M.~Schr{\"o}der}, \bibinfo{author}{K.-H. K{\"u}fer},
\newblock \bibinfo{title}{A constraint-based approach for the two-dimensional rectangular packing problem with orthogonal orientations},
\newblock in: \bibinfo{editor}{B.~Fleischmann}, \bibinfo{editor}{K.-H. Borgwardt}, \bibinfo{editor}{R.~Klein}, \bibinfo{editor}{A.~Tuma} (Eds.), \bibinfo{booktitle}{Operations Research Proceedings 2008}, \bibinfo{publisher}{Springer Berlin Heidelberg}, \bibinfo{address}{Berlin, Heidelberg}, \bibinfo{year}{2009}, pp. \bibinfo{pages}{427--432}.
\bibitem[{Korf et~al.(2010)Korf, Moffitt, and Pollack}]{CSP_abs_rela}
\bibinfo{author}{R.~Korf}, \bibinfo{author}{M.~Moffitt}, \bibinfo{author}{M.~Pollack},
\newblock \bibinfo{title}{{Optimal rectangle packing}},
\newblock \bibinfo{journal}{Annals of Operations Research} \bibinfo{volume}{179} (\bibinfo{year}{2010}) \bibinfo{pages}{261--295}. \URLprefix \url{https://ideas.repec.org/a/spr/annopr/v179y2010i1p261-29510.1007-s10479-008-0463-6.html}. \DOIprefix\doi{10.1007/s10479-008-0463-6}.
\bibitem[{Oliveira et~al.(2016)Oliveira, J{\'u}nior, Silva, and Carravilla}]{packing_heur_survey}
\bibinfo{author}{J.~F. Oliveira}, \bibinfo{author}{A.~N. J{\'u}nior}, \bibinfo{author}{E.~Silva}, \bibinfo{author}{M.~A. Carravilla},
\newblock \bibinfo{title}{A survey on heuristics for the two-dimensional rectangular strip packing problem},
\newblock \bibinfo{journal}{Pesquisa Operacional} \bibinfo{volume}{36} (\bibinfo{year}{2016}) \bibinfo{pages}{197--226}.
\bibitem[{Hopper and Turton(1999)}]{Hopper1999AGA}
\bibinfo{author}{E.~Hopper}, \bibinfo{author}{B.~Turton},
\newblock \bibinfo{title}{A genetic algorithm for a {2D} industrial packing problem},
\newblock \bibinfo{journal}{Computers \& Industrial Engineering} \bibinfo{volume}{37} (\bibinfo{year}{1999}) \bibinfo{pages}{375--378}. \URLprefix \url{https://www.sciencedirect.com/science/article/pii/S0360835299000972}. \DOIprefix\doi{https://doi.org/10.1016/S0360-8352(99)00097-2}.
\bibitem[{Crainic et~al.(2008)Crainic, Perboli, and Tadei}]{Crainic2008ExtremePH}
\bibinfo{author}{T.~G. Crainic}, \bibinfo{author}{G.~Perboli}, \bibinfo{author}{R.~Tadei},
\newblock \bibinfo{title}{Extreme point-based heuristics for three-dimensional bin packing},
\newblock \bibinfo{journal}{INFORMS J. Comput.} \bibinfo{volume}{20} (\bibinfo{year}{2008}) \bibinfo{pages}{368--384}. \URLprefix \url{https://api.semanticscholar.org/CorpusID:2742062}.
\bibitem[{Leung et~al.(2012)Leung, Zhang, Zhou, and Wu}]{SA_hybrid_packing}
\bibinfo{author}{S.~C. Leung}, \bibinfo{author}{D.~Zhang}, \bibinfo{author}{C.~Zhou}, \bibinfo{author}{T.~Wu},
\newblock \bibinfo{title}{A hybrid simulated annealing metaheuristic algorithm for the two-dimensional knapsack packing problem},
\newblock \bibinfo{journal}{Computers \& Operations Research} \bibinfo{volume}{39} (\bibinfo{year}{2012}) \bibinfo{pages}{64--73}. \URLprefix \url{https://www.sciencedirect.com/science/article/pii/S0305054810002510}. \DOIprefix\doi{https://doi.org/10.1016/j.cor.2010.10.022}.
\bibitem[{Alvarez-Valdes et~al.(2008)Alvarez-Valdes, Parreño, and Tamarit}]{grasp_packing}
\bibinfo{author}{R.~Alvarez-Valdes}, \bibinfo{author}{F.~Parreño}, \bibinfo{author}{J.~Tamarit},
\newblock \bibinfo{title}{Reactive {GRASP} for the strip-packing problem},
\newblock \bibinfo{journal}{Computers \& Operations Research} \bibinfo{volume}{35} (\bibinfo{year}{2008}) \bibinfo{pages}{1065--1083}. \URLprefix \url{https://www.sciencedirect.com/science/article/pii/S030505480600150X}. \DOIprefix\doi{https://doi.org/10.1016/j.cor.2006.07.004}.
\bibitem[{Xie and Sahinidis(2008)}]{fac2}
\bibinfo{author}{W.~Xie}, \bibinfo{author}{N.~V. Sahinidis},
\newblock \bibinfo{title}{A branch-and-bound algorithm for the continuous facility layout problem},
\newblock \bibinfo{journal}{{Computers \& Chemical Engineering}} \bibinfo{volume}{32} (\bibinfo{year}{2008}) \bibinfo{pages}{1016--1028}. \URLprefix \url{https://www.sciencedirect.com/science/article/pii/S0098135407001354}. \DOIprefix\doi{https://doi.org/10.1016/j.compchemeng.2007.05.003}.
\bibitem[{Klausnitzer and Lasch(2019)}]{opt_fac_mip}
\bibinfo{author}{A.~Klausnitzer}, \bibinfo{author}{R.~Lasch},
\newblock \bibinfo{title}{Optimal facility layout and material handling network design},
\newblock \bibinfo{journal}{Computers \& Operations Research} \bibinfo{volume}{103} (\bibinfo{year}{2019}) \bibinfo{pages}{237--251}. \URLprefix \url{https://www.sciencedirect.com/science/article/pii/S0305054818302879}. \DOIprefix\doi{https://doi.org/10.1016/j.cor.2018.11.002}.
\bibitem[{Kubal{\'i}k et~al.(2019)Kubal{\'i}k, Kadera, Jirkovsk{\'y}, Kurilla, and Prokop}]{kubalik}
\bibinfo{author}{J.~Kubal{\'i}k}, \bibinfo{author}{P.~Kadera}, \bibinfo{author}{V.~Jirkovsk{\'y}}, \bibinfo{author}{L.~Kurilla}, \bibinfo{author}{{\v{S}}.~Prokop},
\newblock \bibinfo{title}{Plant layout optimization using evolutionary algorithms},
\newblock in: \bibinfo{booktitle}{Industrial Applications of Holonic and Multi-Agent Systems}, \bibinfo{publisher}{Springer International Publishing}, \bibinfo{address}{Cham}, \bibinfo{year}{2019}, pp. \bibinfo{pages}{173--188}.
\bibitem[{Kubalík et~al.(2023)Kubalík, Kurilla, and Kadera}]{kubalik2}
\bibinfo{author}{J.~Kubalík}, \bibinfo{author}{L.~Kurilla}, \bibinfo{author}{P.~Kadera},
\newblock \bibinfo{title}{Facility layout problem with alternative facility variants},
\newblock \bibinfo{journal}{Applied Sciences} \bibinfo{volume}{13} (\bibinfo{year}{2023}). \DOIprefix\doi{10.3390/app13085032}.
\bibitem[{Xiao et~al.(2017)Xiao, Xie, Kulturel-Konak, and Konak}]{fac_mip_but_nonexact}
\bibinfo{author}{Y.~Xiao}, \bibinfo{author}{Y.~Xie}, \bibinfo{author}{S.~Kulturel-Konak}, \bibinfo{author}{A.~Konak},
\newblock \bibinfo{title}{A problem evolution algorithm with linear programming for the dynamic facility layout problem—a general layout formulation},
\newblock \bibinfo{journal}{Computers \& Operations Research} \bibinfo{volume}{88} (\bibinfo{year}{2017}) \bibinfo{pages}{187--207}. \URLprefix \url{https://www.sciencedirect.com/science/article/pii/S0305054817301648}. \DOIprefix\doi{https://doi.org/10.1016/j.cor.2017.06.025}.
\bibitem[{Grus. et~al.(2023)Grus., Hanzálek., Barri., and Vacula.}]{icores23}
\bibinfo{author}{J.~Grus.}, \bibinfo{author}{Z.~Hanzálek.}, \bibinfo{author}{D.~Barri.}, \bibinfo{author}{P.~Vacula.},
\newblock \bibinfo{title}{Automatic placer for analog circuits using integer linear programming warm started by graph drawing},
\newblock in: \bibinfo{booktitle}{Proceedings of the 12th International Conference on Operations Research and Enterprise Systems - ICORES,}, \bibinfo{organization}{INSTICC}, \bibinfo{publisher}{SciTePress}, \bibinfo{year}{2023}, pp. \bibinfo{pages}{106--116}. \DOIprefix\doi{10.5220/0011789300003396}.
\bibitem[{Grus and Hanz{\'a}lek(2024)}]{icores2023-rozsireni}
\bibinfo{author}{J.~Grus}, \bibinfo{author}{Z.~Hanz{\'a}lek},
\newblock \bibinfo{title}{Matheuristic local search for the placement of analog integrated circuits},
\newblock in: \bibinfo{editor}{F.~Liberatore}, \bibinfo{editor}{S.~Wesolkowski}, \bibinfo{editor}{M.~Demange}, \bibinfo{editor}{G.~H. Parlier} (Eds.), \bibinfo{booktitle}{Operations Research and Enterprise Systems}, \bibinfo{publisher}{Springer Nature Switzerland}, \bibinfo{address}{Cham}, \bibinfo{year}{2024}, pp. \bibinfo{pages}{178--200}.
\bibitem[{Della~Croce and Scatamacchia(2020)}]{lpt}
\bibinfo{author}{F.~Della~Croce}, \bibinfo{author}{R.~Scatamacchia},
\newblock \bibinfo{title}{The longest processing time rule for identical parallel machines revisited},
\newblock \bibinfo{journal}{Journal of Scheduling} \bibinfo{volume}{23} (\bibinfo{year}{2020}) \bibinfo{pages}{163--176}. \URLprefix \url{https://doi.org/10.1007/s10951-018-0597-6}. \DOIprefix\doi{10.1007/s10951-018-0597-6}.
\bibitem[{Camm et~al.(1990)Camm, Raturi, and Tsubakitani}]{bigM}
\bibinfo{author}{J.~D. Camm}, \bibinfo{author}{A.~S. Raturi}, \bibinfo{author}{S.~Tsubakitani},
\newblock \bibinfo{title}{{Cutting Big M down to Size}},
\newblock \bibinfo{journal}{Interfaces} \bibinfo{volume}{20} (\bibinfo{year}{1990}) \bibinfo{pages}{61--66}. \URLprefix \url{http://www.jstor.org/stable/25061401}.
\bibitem[{Mendes et~al.(2009)Mendes, Gonçalves, and Resende}]{ga_rcpsp}
\bibinfo{author}{J.~Mendes}, \bibinfo{author}{J.~Gonçalves}, \bibinfo{author}{M.~Resende},
\newblock \bibinfo{title}{A random key based genetic algorithm for the resource constrained project scheduling problem},
\newblock \bibinfo{journal}{Computers \& Operations Research} \bibinfo{volume}{36} (\bibinfo{year}{2009}) \bibinfo{pages}{92--109}. \URLprefix \url{https://www.sciencedirect.com/science/article/pii/S0305054807001359}. \DOIprefix\doi{https://doi.org/10.1016/j.cor.2007.07.001}.
\bibitem[{Chazelle(1983)}]{blf_quadratic}
\bibinfo{author}{B.~Chazelle},
\newblock \bibinfo{title}{The bottom-left bin-packing heuristic: An efficient implementation},
\newblock \bibinfo{journal}{IEEE Transactions on Computers} \bibinfo{volume}{C-32} (\bibinfo{year}{1983}) \bibinfo{pages}{697--707}. \DOIprefix\doi{10.1109/TC.1983.1676307}.
\bibitem[{Hansen and Ostermeier(2001)}]{hansen_cma}
\bibinfo{author}{N.~Hansen}, \bibinfo{author}{A.~Ostermeier},
\newblock \bibinfo{title}{Completely derandomized self-adaptation in evolution strategies},
\newblock \bibinfo{journal}{Evolutionary Computation} \bibinfo{volume}{9} (\bibinfo{year}{2001}) \bibinfo{pages}{159--195}. \DOIprefix\doi{10.1162/106365601750190398}.
\bibitem[{Hansen et~al.(2010)Hansen, Auger, Ros, Finck, and Po\v{s}\'{\i}k}]{meta_comp}
\bibinfo{author}{N.~Hansen}, \bibinfo{author}{A.~Auger}, \bibinfo{author}{R.~Ros}, \bibinfo{author}{S.~Finck}, \bibinfo{author}{P.~Po\v{s}\'{\i}k},
\newblock \bibinfo{title}{Comparing results of 31 algorithms from the black-box optimization benchmarking {BBOB}-2009},
\newblock in: \bibinfo{booktitle}{Proceedings of the 12th Annual Conference Companion on Genetic and Evolutionary Computation}, GECCO '10, \bibinfo{publisher}{Association for Computing Machinery}, \bibinfo{address}{New York, NY, USA}, \bibinfo{year}{2010}, p. \bibinfo{pages}{1689–1696}. \URLprefix \url{https://doi.org/10.1145/1830761.1830790}. \DOIprefix\doi{10.1145/1830761.1830790}.
\bibitem[{Eiben et~al.(1999)Eiben, Hinterding, and Michalewicz}]{param_control_overview}
\bibinfo{author}{A.~Eiben}, \bibinfo{author}{R.~Hinterding}, \bibinfo{author}{Z.~Michalewicz},
\newblock \bibinfo{title}{Parameter control in evolutionary algorithms},
\newblock \bibinfo{journal}{IEEE Transactions on Evolutionary Computation} \bibinfo{volume}{3} (\bibinfo{year}{1999}) \bibinfo{pages}{124--141}. \DOIprefix\doi{10.1109/4235.771166}.
\bibitem[{Sakurai et~al.(2010)Sakurai, Takada, Kawabe, and Tsuruta}]{rl_basic}
\bibinfo{author}{Y.~Sakurai}, \bibinfo{author}{K.~Takada}, \bibinfo{author}{T.~Kawabe}, \bibinfo{author}{S.~Tsuruta},
\newblock \bibinfo{title}{A method to control parameters of evolutionary algorithms by using reinforcement learning},
\newblock in: \bibinfo{booktitle}{2010 Sixth International Conference on Signal-Image Technology and Internet Based Systems}, \bibinfo{year}{2010}, pp. \bibinfo{pages}{74--79}. \DOIprefix\doi{10.1109/SITIS.2010.22}.
\bibitem[{Sharma et~al.(2019)Sharma, Komninos, L\'{o}pez-Ib\'{a}\~{n}ez, and Kazakov}]{dqnga}
\bibinfo{author}{M.~Sharma}, \bibinfo{author}{A.~Komninos}, \bibinfo{author}{M.~L\'{o}pez-Ib\'{a}\~{n}ez}, \bibinfo{author}{D.~Kazakov},
\newblock \bibinfo{title}{Deep reinforcement learning based parameter control in differential evolution},
\newblock in: \bibinfo{booktitle}{Proceedings of the Genetic and Evolutionary Computation Conference}, GECCO '19, \bibinfo{publisher}{Association for Computing Machinery}, \bibinfo{address}{New York, NY, USA}, \bibinfo{year}{2019}, p. \bibinfo{pages}{709–717}. \URLprefix \url{https://doi.org/10.1145/3321707.3321813}. \DOIprefix\doi{10.1145/3321707.3321813}.
\bibitem[{Chapman and Lechner(2020)}]{chapman_lechner}
\bibinfo{author}{J.~Chapman}, \bibinfo{author}{M.~Lechner}, \bibinfo{title}{{Keras Documentation: Deep Q-learning for atari breakout}}, \bibinfo{howpublished}{\url{https://keras.io/examples/rl/deep_q_network_breakout/}}, \bibinfo{year}{2020}. \bibinfo{note}{Accessed: 2023-07-19}.
\bibitem[{{Gurobi Optimization, LLC}(2021)}]{gurobi}
\bibinfo{author}{{Gurobi Optimization, LLC}}, \bibinfo{title}{{Gurobi Optimizer Reference Manual}}, \bibinfo{howpublished}{\url{https://www.gurobi.com}}, \bibinfo{year}{2021}. \bibinfo{note}{Accessed: 2023-07-19}.
\bibitem[{Abadi et~al.(2015)Abadi, Agarwal, Barham, Brevdo, Chen, Citro, Corrado, Davis, Dean, Devin, Ghemawat, Goodfellow, Harp, Irving, Isard, Jia, Jozefowicz, Kaiser, Kudlur, Levenberg, Man\'{e}, Monga, Moore, Murray, Olah, Schuster, Shlens, Steiner, Sutskever, Talwar, Tucker, Vanhoucke, Vasudevan, Vi\'{e}gas, Vinyals, Warden, Wattenberg, Wicke, Yu, and Zheng}]{tensorflow2015-whitepaper}
\bibinfo{author}{M.~Abadi}, \bibinfo{author}{A.~Agarwal}, \bibinfo{author}{P.~Barham}, \bibinfo{author}{E.~Brevdo}, \bibinfo{author}{Z.~Chen}, \bibinfo{author}{C.~Citro}, \bibinfo{author}{G.~S. Corrado}, \bibinfo{author}{A.~Davis}, \bibinfo{author}{J.~Dean}, \bibinfo{author}{M.~Devin}, \bibinfo{author}{S.~Ghemawat}, \bibinfo{author}{I.~Goodfellow}, \bibinfo{author}{A.~Harp}, \bibinfo{author}{G.~Irving}, \bibinfo{author}{M.~Isard}, \bibinfo{author}{Y.~Jia}, \bibinfo{author}{R.~Jozefowicz}, \bibinfo{author}{L.~Kaiser}, \bibinfo{author}{M.~Kudlur}, \bibinfo{author}{J.~Levenberg}, \bibinfo{author}{D.~Man\'{e}}, \bibinfo{author}{R.~Monga}, \bibinfo{author}{S.~Moore}, \bibinfo{author}{D.~Murray}, \bibinfo{author}{C.~Olah}, \bibinfo{author}{M.~Schuster}, \bibinfo{author}{J.~Shlens}, \bibinfo{author}{B.~Steiner}, \bibinfo{author}{I.~Sutskever}, \bibinfo{author}{K.~Talwar}, \bibinfo{author}{P.~Tucker}, \bibinfo{author}{V.~Vanhoucke}, \bibinfo{author}{V.~Vasudevan}, \bibinfo{author}{F.~Vi\'{e}gas},
  \bibinfo{author}{O.~Vinyals}, \bibinfo{author}{P.~Warden}, \bibinfo{author}{M.~Wattenberg}, \bibinfo{author}{M.~Wicke}, \bibinfo{author}{Y.~Yu}, \bibinfo{author}{X.~Zheng}, \bibinfo{title}{{TensorFlow}: Large-scale machine learning on heterogeneous systems}, \bibinfo{year}{2015}. \URLprefix \url{https://www.tensorflow.org/}, \bibinfo{note}{accessed: 2023-07-19}.
\bibitem[{Hansen et~al.(2019)Hansen, Akimoto, and Baudis}]{hansen2019pycma}
\bibinfo{author}{N.~Hansen}, \bibinfo{author}{Y.~Akimoto}, \bibinfo{author}{P.~Baudis}, \bibinfo{title}{{CMA-ES/pycma} on {G}ithub}, \bibinfo{howpublished}{Zenodo, DOI:10.5281/zenodo.2559634}, \bibinfo{year}{2019}. \URLprefix \url{https://doi.org/10.5281/zenodo.2559634}. \DOIprefix\doi{10.5281/zenodo.2559634}, \bibinfo{note}{accessed: 2023-07-19}.
\bibitem[{Marsili-Libelli and Alba(2000)}]{adaptivegenetic}
\bibinfo{author}{S.~Marsili-Libelli}, \bibinfo{author}{P.~Alba},
\newblock \bibinfo{title}{Adaptive mutation in genetic algorithms},
\newblock \bibinfo{journal}{Soft Computing} \bibinfo{volume}{4} (\bibinfo{year}{2000}) \bibinfo{pages}{76--80}. \DOIprefix\doi{10.1007/s005000000042}.
\bibitem[{Lin(2009)}]{adaptivegenetic2}
\bibinfo{author}{C.~Lin},
\newblock \bibinfo{title}{An adaptive genetic algorithm based on population diversity strategy},
\newblock in: \bibinfo{booktitle}{2009 Third International Conference on Genetic and Evolutionary Computing}, \bibinfo{year}{2009}, pp. \bibinfo{pages}{93--96}. \DOIprefix\doi{10.1109/WGEC.2009.67}.
\bibitem[{Chen et~al.(2007)Chen, Lin, Wang, and Cheng}]{gsrc_area_only}
\bibinfo{author}{D.-S. Chen}, \bibinfo{author}{C.-T. Lin}, \bibinfo{author}{Y.-W. Wang}, \bibinfo{author}{C.-H. Cheng},
\newblock \bibinfo{title}{Fixed-outline floorplanning using robust evolutionary search},
\newblock \bibinfo{journal}{Engineering Applications of Artificial Intelligence} \bibinfo{volume}{20} (\bibinfo{year}{2007}) \bibinfo{pages}{821--830}. \URLprefix \url{https://www.sciencedirect.com/science/article/pii/S0952197606001643}. \DOIprefix\doi{https://doi.org/10.1016/j.engappai.2006.10.006}.
\bibitem[{Manikas(2012)}]{mcnc_bench}
\bibinfo{author}{T.~W. Manikas}, \bibinfo{title}{{MCNC} benchmark netlists for floorplanning and placement}, \bibinfo{year}{2012}. \URLprefix \url{https://s2.smu.edu/~manikas/Benchmarks/MCNC_Benchmark_Netlists.html}, \bibinfo{note}{accessed: 2023-03-14}.

\end{thebibliography}
\end{document}